\documentclass{article}

\usepackage{arxiv}
\usepackage{authblk}

\usepackage{datetime2} 
\usepackage{fancyhdr}
\fancypagestyle{firstpagestyle}{
    \fancyhf{}
    \renewcommand{\headrulewidth}{0pt}
}

\usepackage{fullpage} 
\usepackage{wrapfig}
\usepackage{multirow}
\usepackage[utf8]{inputenc}
\usepackage[english]{babel}
\usepackage{amssymb,amsfonts}
\usepackage{amsthm}
\usepackage{float}
\usepackage{booktabs}
\usepackage{color}
\usepackage{listings}
\usepackage{rotating}
\usepackage[titletoc]{appendix}
\usepackage{minitoc}
\usepackage{pifont}
\usepackage{stmaryrd}
\usepackage{array}
\usepackage[algoruled]{algorithm2e}

\usepackage{graphicx}
\usepackage{multicol}
\usepackage{enumitem}
\usepackage{framed}
\usepackage[
	operators,
	sets,
	landau,
	probability,
	notions,
	logic,
	ff,
	mm,
	advantage,
	events,
	complexity,
	asymptotics,
	keys,
	lambda]{cryptocode}
\usepackage{comment}
\usepackage{adjustbox}
\usepackage{varioref}
\usepackage{hyperref}
\usepackage{breakcites}
\usepackage[section]{placeins}
\usepackage{subcaption}
\usepackage{tcolorbox}

\usepackage{tikz}
\usetikzlibrary{arrows.meta,decorations.pathmorphing,decorations.footprints,fadings,calc,trees,mindmap,shadows,decorations.text,patterns,positioning,shapes,matrix,fit}

\usepackage{url} 
\usepackage{framed}
\usepackage{comment}
\usepackage{adjustbox}
\usepackage[nameinlink,capitalize]{cleveref}

\usepackage{pgf, tikz}
\usetikzlibrary{arrows, automata, positioning}

\floatstyle{boxed}
\definecolor{cadmiumorange}{rgb}{0.93, 0.53, 0.18}
\definecolor{emerald}{rgb}{0.31, 0.78, 0.47}
\definecolor{amaranth}{rgb}{0.9, 0.17, 0.31}
\definecolor{candypink}{rgb}{0.89, 0.44, 0.48}
\definecolor{caribbeangreen}{rgb}{0.0, 0.8, 0.6}
\definecolor{cornflowerblue}{rgb}{0.39, 0.58, 0.93}
\definecolor{limegreen}{rgb}{0.2, 0.8, 0.2}
\definecolor{mayablue}{rgb}{0.21,0.49,0.74}

\usepackage{xcolor}

\colorlet{blue}{mayablue}
\definecolor{green}{HTML}{2a9d8f}
\colorlet{orange}{cadmiumorange}

\hypersetup{
  linkcolor  = amaranth,
    filecolor=magenta,      
    urlcolor=teal,
    citecolor=mayablue, 
    colorlinks = true,
    breaklinks = true,
    bookmarksopen=true
} 

\usepackage{epigraph}
\usepackage{natbib}

\NewDocumentCommand{\hl}{ mO{} }{\textcolor{mayablue}{\textsuperscript{\textit{Hanlin}}\textsf{\textbf{\small[#1]}}}}
\NewDocumentCommand{\todo}{ mO{} }{\textcolor{mayablue}{\textsuperscript{\textit{Hanlin}}\textsf\textbf{\small[{TODO: #1]}}}}

\newtheorem{corollary}{Corollary}

\newtheorem{proposition}{Proposition}

\definecolor{thmref}{HTML}{2a9d8f}
\crefname{assumption}{\textcolor{thmref}{Assumption}}{\textcolor{thmref}{Assumptions}}
\Crefname{assumption}{\textcolor{thmref}{Assumption}}{\textcolor{thmref}{Assumptions}}
\crefname{definition}{\textcolor{thmref}{Definition}}{\textcolor{thmref}{Definitions}}
\Crefname{definition}{\textcolor{thmref}{Definition}}{\textcolor{thmref}{Definitions}}
\crefname{theorem}{\textcolor{thmref}{Theorem}}{\textcolor{thmref}{Theorems}}
\Crefname{theorem}{\textcolor{thmref}{Theorem}}{\textcolor{thmref}{Theorems}}
\crefname{lemma}{\textcolor{thmref}{Lemma}}{\textcolor{thmref}{Lemmas}}
\Crefname{lemma}{\textcolor{thmref}{Lemma}}{\textcolor{thmref}{Lemmas}}
\crefname{corollary}{\textcolor{thmref}{Corollary}}{\textcolor{thmref}{Corollaries}}
\Crefname{corollary}{\textcolor{thmref}{Corollary}}{\textcolor{thmref}{Corollaries}}
\crefname{proposition}{\textcolor{thmref}{Proposition}}{\textcolor{thmref}{Propositions}}
\Crefname{proposition}{\textcolor{thmref}{Proposition}}{\textcolor{thmref}{Propositions}}
\crefname{remark}{\textcolor{thmref}{Remark}}{\textcolor{thmref}{Remarks}}
\Crefname{remark}{\textcolor{thmref}{Remark}}{\textcolor{thmref}{Remarks}}
\creflabelformat{assumption}{#2\textcolor{thmref}{#1}#3}
\creflabelformat{definition}{#2\textcolor{thmref}{#1}#3}
\creflabelformat{theorem}{#2\textcolor{thmref}{#1}#3}
\creflabelformat{lemma}{#2\textcolor{thmref}{#1}#3}
\creflabelformat{corollary}{#2\textcolor{thmref}{#1}#3}
\creflabelformat{proposition}{#2\textcolor{thmref}{#1}#3}
\creflabelformat{remark}{#2\textcolor{thmref}{#1}#3}

\usepackage{tabularx}
\graphicspath{{figures/}}

\definecolor{mayablue}{rgb}{0.21,0.49,0.74}
\definecolor{amaranth}{rgb}{0.9,0.17,0.31}
\definecolor{openpurple}{RGB}{112,48,160}
\definecolor{NFrozen}{HTML}{02507D}
\definecolor{NTrain}{HTML}{A66A00}
\definecolor{NSup}{HTML}{016B4E}
\newcommand{\cf}[1]{{\color{NFrozen}#1}}
\newcommand{\ct}[1]{{\color{NTrain}#1}}
\newcommand{\cs}[1]{{\color{NSup}#1}}

\providecommand{\todo}[1]{{\color{openpurple}\small\textsf{Open.} #1}}

\definecolor{editblue}{RGB}{0,90,200}
\newcommand{\edit}[1]{#1}

\makeatletter
\def\thm@space@setup{\thm@preskip=6pt \thm@postskip=6pt}
\makeatother
\makeatletter
\@ifundefined{proposition}{\newtheorem{proposition}{Proposition}}{}
\@ifundefined{corollary}{\newtheorem{corollary}{Corollary}}{}
\makeatother

\author[1,2]{Yueying Li\thanks{Work done outside of NVIDIA, while conducting research at MIT and Cornell.}}
\author[3]{Jingxuan Fan}
\author[4]{Kiant\'e Brantley}
\author[4]{Sham M. Kakade}
\author[4]{Hanlin Zhang}
\affil[1]{Massachusetts Institute of Technology}
\affil[2]{Cornell University}
\affil[3]{Scale AI}
\affil[4]{Harvard University}
\affil[ ]{\normalsize\textit{\today}}

\title{Unsupervised Multimodal Reward Modeling}

\begin{document}
\addtocontents{toc}{\protect\setcounter{tocdepth}{-1}}
\maketitle
\thispagestyle{firstpagestyle}
\setlist[enumerate]{leftmargin=*}
\setlist[itemize]{leftmargin=*}
\setlist[description]{leftmargin=0pt}

\begin{abstract}
    Vision-language reward models are trained on annotated preference pairs, which are expensive to collect and can disagree with each other, and are ranked by benchmark accuracy. We build one with no human preference label: a linear head on an intermediate layer of a frozen Qwen3-VL backbone, trained to score a web caption above a perturbed version of the same caption; labeled data enters only to select the layer. The layer is the decisive choice, worth 5.8 points over the last layer, and one recipe transferred without re-tuning beats or at least matches GPT-4o on three benchmarks at 32B, 6.6 points ahead on VL-RewardBench, from a few thousand parameters at about a thousandth of fine-tuning compute. At an equal pair budget, perturbations teach a frozen head 3 to 5 points more than curated RLAIF-V pairs. Because the backbone is frozen the objective is convex and the perturbation sampler specifies the reward, which the benchmark does not identify: two heads trained on disjoint halves of the same perturbations tie on the benchmark and differ by 6.4 points in hallucination judging. Retraining the head on the policy's own verifier-labelled candidates cuts caption hallucination by 5.1 CHAIR points and lifts image-generation pass rate by 5.4, while scoring lower on every reward benchmark. Together, an intermediate-layer probe, free same-image perturbations and a verifier-driven bootstrap turn give a recipe for multimodal reward modeling that needs no annotator, costs a fraction of fine-tuning, and is specified by the comparisons it is trained on rather than by a benchmark rank.
\end{abstract}

\begin{figure*}[!h]
    \centering
    \includegraphics[width=\linewidth]{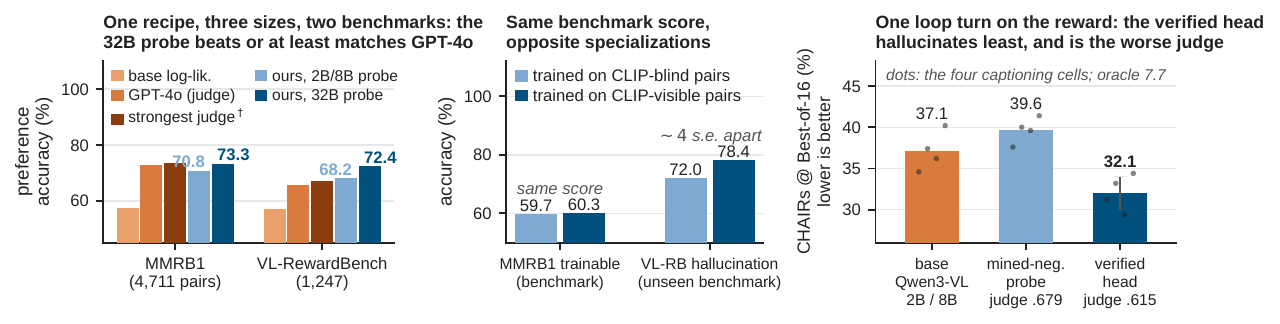}
    \vskip -0.05in
    \caption{\textbf{Reward modeling with no human preference label.} \textbf{(a)} A linear probe on a frozen
Qwen3-VL backbone, trained on caption perturbations, beats or at least matches GPT-4o at 32B. \textbf{(b)} Two
probes trained on disjoint halves of the same data tie on the benchmark and differ by 6.4 points on
hallucination judging: the sampler, not the score, sets the reward. \textbf{(c)} Retraining the head on the
policy's own verifier-labelled candidates cuts caption hallucination 5.1 CHAIR points below the base generators
(\cref{sec:midtraining}).}
    \label{fig:teaser}
\vskip -0.15in
\end{figure*}

\section{Introduction}
\label{sec:intro}

Reinforcement learning from human feedback \citep{christiano2017deeprlhf,ouyang2022instructgpt} has
made the reward model the channel through which preferences enter post-training. For
vision-language models the same design is now standard: a scalar scorer over an image, a prompt and a
response, trained on annotated preference pairs and ranked on preference benchmarks
\citep{yasunaga2025mmrewardbench,li2024vlrewardbench,hu2025mmrb2}. The design has two costs: multimodal
preference pairs are expensive to annotate and the annotations disagree with each other
\citep{yu2024rlaifv}. Once the model is built, benchmark accuracy is how it is compared, selected and trusted.
VL-RewardBench accuracy correlates with Best-of-$N$ accuracy on MMMU-Pro for generative judges
(Pearson's $r>0.9$; \citealp{li2024vlrewardbench}); what sets that number for a scalar reward model,
and when it stops predicting deployment, is the question this paper takes up.

\begin{figure*}[t]
    \centering
    \includegraphics[width=0.82\linewidth]{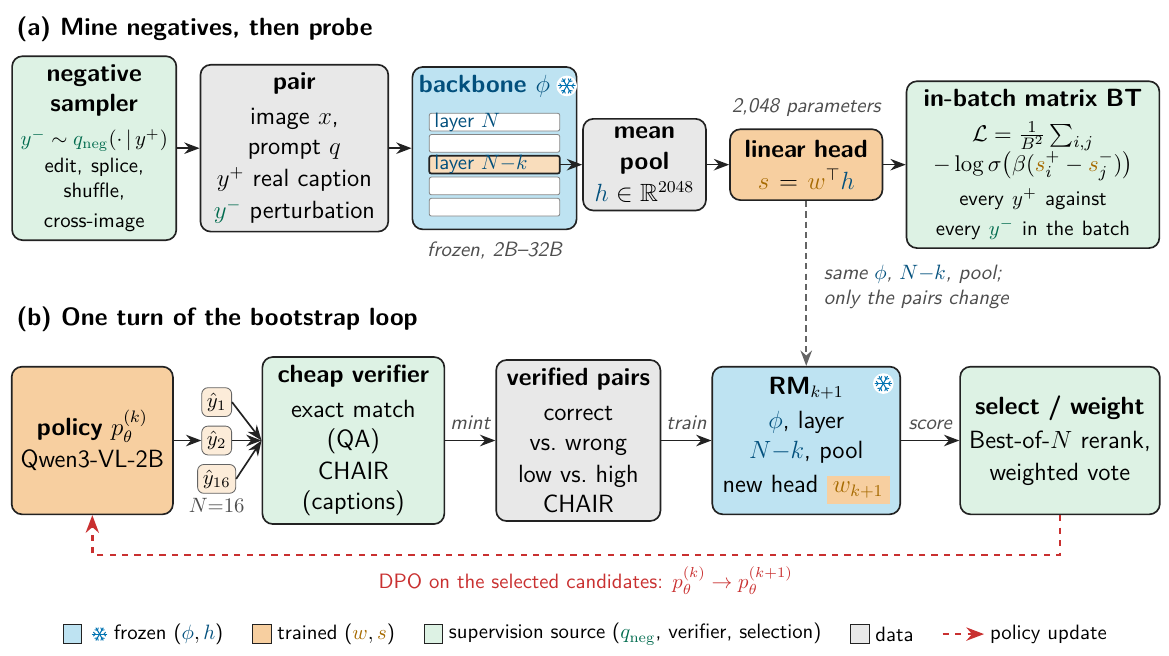}
    \caption{\textbf{(a) The reward from a linear probe.} A pair (image, prompt, real caption,
    same-image perturbation) passes once through the frozen backbone $\cf{\phi}$; the layer-$\cf{N{-}k}$
    states are mean-pooled over response tokens into $\cf{h}$ and scored by a bias-free head $\ct{w}$,
    trained with in-batch matrix Bradley--Terry; the sampler $\cs{q_{\mathrm{neg}}}$ is the only place
    supervision enters. \textbf{(b) One turn of the bootstrap loop} (\cref{sec:midtraining}): cheap
    verifiers label the policy's own Best-of-$N$ candidates and the same head is retrained on the minted
    pairs. \cf{Blue} frozen, \ct{orange} trained, \cs{green} supervision.}
    \label{fig:method}
\vskip -0.1in
\end{figure*}

Text reward models show that a reward is a comparison before it is a number: \edit{Bradley--Terry training identifies a reward only up to a shift that may differ across prompts \citep{bradley1952rank,rafailov2023dpo}, so} a score of .8 has no stable meaning across prompts, candidate pools or checkpoints. The scalar hides the relative preference that was always the operative signal \citep{\edit{sun2025rethinking,}zhang2026rlcd}, so what a reward model
learns is set by which comparisons it is trained on. And those comparisons need no annotator: web continuations as chosen against in-batch continuations
as rejected train a text reward model with no label at all \citep{fan2026rbs}. We take both findings to the vision-language setting, make the
comparison the object of study, and remove the human labels entirely. The reward model is a linear
head on the hidden states of an intermediate layer of a frozen backbone, trained by in-batch Bradley--Terry
\citep{bradley1952rank} against same-image perturbations of web captions or same-seed re-renders of a perturbed prompt
(\cref{fig:method}, \cref{sec:methods}). \edit{The claim is about the training
data: the probe layer is selected on one labeled benchmark, which we therefore treat as development and
never as transfer evidence, and the loop of \cref{sec:midtraining} uses verifier labels, not annotators.} Two things then decide what the head learns: the representation it reads, which sets the signal
a linear head can extract, and the comparison distribution it is trained on, which sets the direction
of the reward. Every choice in the paper is one of the two, and the verified policy errors of
\cref{sec:midtraining} are the second variable's final instance. A head of a few thousand parameters
built this way beats GPT-4o by 6.6 points on VL-RewardBench at 32B, for about a thousandth of the
compute of full backbone fine-tuning, and is level with GPT-4o on Multimodal RewardBench and MMRB2. At
an equal pair count, the mined pairs teach a frozen head more than curated RLAIF-V pairs do.

Because the backbone is frozen, the objective is transparent enough to say what that number is. It is
ranking noise-contrastive estimation over fixed features \citep{gutmann2010nce,ma2018rankingnce}, so with a fixed quadratic penalty it is convex with a unique minimizer, characterized as an average of the negative sampler's
difference vectors weighted by how badly the head still misorders each pair (\cref{sec:identification}). AdamW does not optimize that fixed objective, so where the trained heads land is measured rather than derived. In our runs no training-side change
(pool size, steps, curriculum, loss form or optimizer setting) moves the benchmark by more than .014, so the representation and the
sampler are the two choices that matter. The representation is the
backbone's, and our findings about it hold across backbones. On four Qwen3-VL sizes and on a second
family, GLM-4.1V-9B, the best probe layer sits well below the top, twelve layers down for 8B, 32B
and GLM-4.1V-9B. The same recipe with no re-tuning lifts each of them above its own untrained
judge (\cref{sec:generalization}). The sampler is the other choice, and it is where the benchmark stops
being informative.

\edit{Which pairs train the head decides which reward it learns.} Two probes trained on disjoint subsets of the same perturbations, split by whether CLIP
can see the edit, reach similar benchmark accuracy and differ by 6.4 points in hallucination judging (\cref{fig:teaser}b). Deployed for Best-of-$N$ reranking, benchmark rank orders utility in neither direction, and where the
probe does beat majority vote, by 7 to 11 points, a training-free log-likelihood judge matches it. The analysis leaves one sampler to try: the policy's own errors. We mint those labels
with cheap verifiers on Best-of-$N$ candidates, for captions and for generated images; the resulting
head cuts caption hallucination by 5.1 CHAIR points below the base Qwen3-VL-2B and 8B generators, while its VL-RewardBench accuracy falls below that of the probe it replaced (\cref{fig:teaser}c). \edit{One DPO \citep{rafailov2023dpo} turn on the policy's own pairs also cuts its hallucination by 3 to 6 CHAIR \citep{rohrbach2018chair} points at no QA cost
(\cref{tab:actor_dpo}).}

Our contributions combine probing, negative sampling and self-improvement, each studied on the
policy side before (\cref{sec:related}), into one reward-side recipe that uses no human preference
label. (i) \emph{Probed, not trained in:} a frozen vision-language backbone with a linear head (2,048 parameters at 2B) at
a depth set by the backbone, at or above GPT-4o on three benchmarks at 32B (\cref{sec:generalization}).
(ii) \emph{Supervision as a sampler:} same-image perturbations written as a mixture of operators that
apply to captions and to rendered images alike, with an argument for which reward direction they select
and an ablation of the operators in both modalities (\cref{sec:identification}, \cref{tab:corruption}).
(iii) \emph{What the score tracks:} benchmark accuracy plateaus under every training-side change we tried, the learned direction is set by the
sampler, and benchmark rank predicts Best-of-$N$ rank in neither direction (\cref{sec:dissociation,sec:bon}).
(iv) \emph{A bootstrap loop} that replaces annotators with cheap verifiers on the policy's own candidates,
the one supervision that moved deployment, shown for captions and for image generation
(\cref{sec:midtraining}).

\newcommand{\figcorruption}{%
\begin{figure}[t]
\centering
\includegraphics[width=\linewidth]{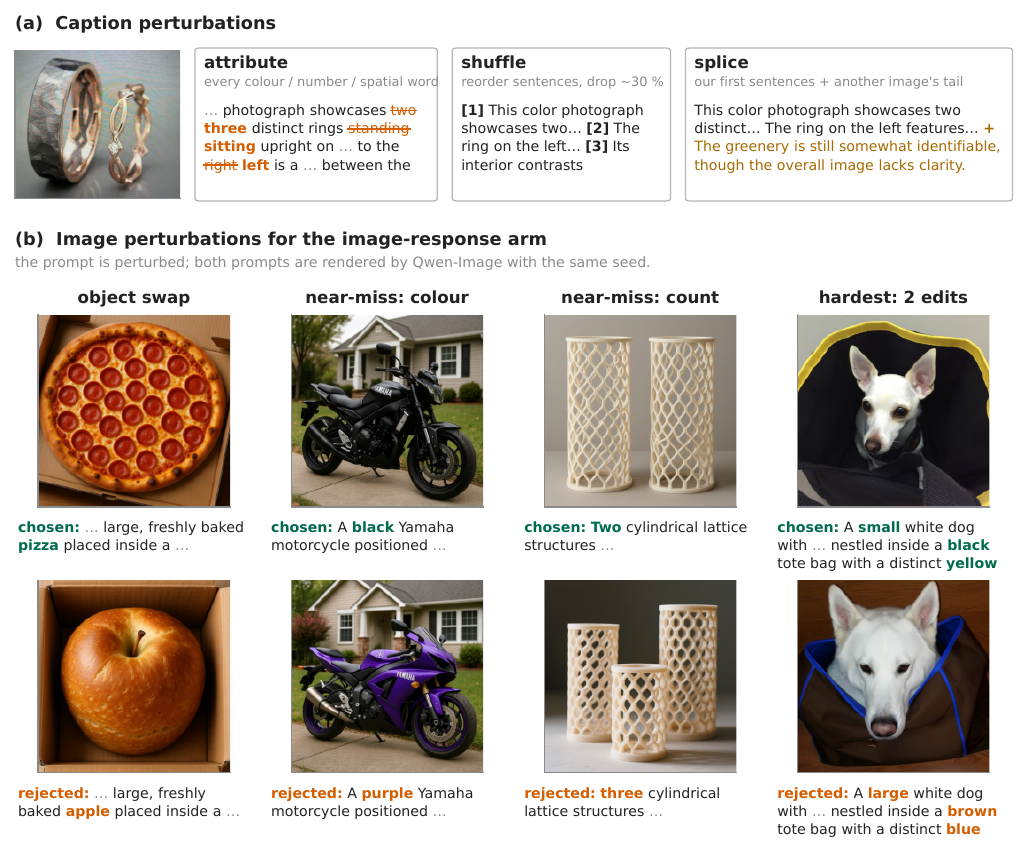}
\caption{Negative mining by perturbation, laid out after ImageNet-C \citep{hendrycks2019imagenetc}. \textbf{(a)} A PixMo image and three operators applied to its
human caption: attribute swaps every colour, number and spatial word; shuffle reorders and drops
sentences; splice continues the true caption with another image's. Red marks the edit, orange the
foreign text. \textbf{(b)} The image-response arm perturbs the prompt instead and renders both prompts
with Qwen-Image under the same seed, so the two prompts differ only in the edited span.}
\label{fig:corruption}
\vskip -0.1in
\end{figure}}

\section{Methodology}
\label{sec:methods}

\subsection{Algorithm}
\label{sec:algorithm}

We score each response with a bias-free linear head on mean-pooled response-token states from
layer $\cf{N{-}k}$ of a frozen Qwen3-VL backbone \citep{qwen3vl2025}. The triple (layer,
pooling, head) is the design space the rest of the paper sweeps (\cref{fig:method}a; colours as in its
caption). For an image $x$, a prompt $q$ and a response
$y$, let $\cf{h^{(N-k)}_t}$ be the hidden state of layer $\cf{N{-}k}$ at response token $t$; the reward
is
\begin{equation}
\label{eq:readout}
\ct{s_w}(x,q,y) \;=\; \ct{w}^{\top}\,\frac{1}{|y|}\sum_{t\in y} \cf{h^{(N-k)}_t} ,
\end{equation}
so the 2B model trains $2{,}048$ parameters and the backbone none. A micro-batch of $B$ preference
pairs $\{(x_i,q_i,y_i^{+},y_i^{-})\}_{i=1}^{B}$ gives the in-batch matrix Bradley--Terry loss, every
chosen response against every rejected one,
\begin{equation}
\label{eq:bt_loss}
\mathcal{L}
= \frac{1}{B^{2}}\sum_{i=1}^{B}\sum_{j=1}^{B}
-\log \sigma\!\big(\beta\,(\ct{s^{+}_{i}} - \ct{s^{-}_{j}})\big),
\end{equation}
with $\beta{=}0.1$ the inverse temperature and $B{=}8$, so each positive meets eight negatives per step
rather than one. Note that supervision enters only through the negative sampler, which is what makes
the sampler the object \cref{sec:identification} reasons about. For a positive $(x,q,y^{+})$ the
rejected response is drawn from a mixture of perturbation operators applied to the positive itself,
\begin{equation}
\label{eq:sampler}
\cs{y^{-}}\;\sim\;\cs{q_{\mathrm{neg}}}(\cdot\mid x,q,y^{+})\;=\;\sum_{o}\pi_{o}\,\cs{\mathcal{T}_{o}}(y^{+};x,q),
\qquad \textstyle\sum_o \pi_o = 1,
\end{equation}
and the modality fixes what the operators act on. Unlike the pixel-level perturbations of ImageNet-C
\citep{hendrycks2019imagenetc}, which perturb the input, these operators edit the \emph{response} and change a
fact while leaving the input untouched. For text responses $\mathcal{T}_{\mathrm{near}}$ substitutes $k$
editable spans of $y^{+}$ (colour, number, size or spatial words), $\mathcal{T}_{\mathrm{splice}}$
replaces its tail with another caption's, $\mathcal{T}_{\mathrm{shuf}}$ permutes its clauses, and
$\mathcal{T}_{\times}$ returns the caption of another image; the recipe uses $\pi_{\times}{=}0.15$ and
$k{=}1$ (\cref{fig:corruption,app:training}). For example, a caption reading \emph{two brown horses beside a red fence} becomes \emph{two brown
horses beside a blue fence}: one colour word moves, the token count is unchanged, and the pair
differs only in a fact about the image. \edit{For image responses the
operator acts on the prompt and a fixed-seed generator $G$ renders both sides, $y^{+}=G(q;s)$ and
$y^{-}=G(\mathcal{T}_{\mathrm{near}}(q);s)$. The usual label-free alternative is to have a model write the rejected response, as in constitutional
and AI-feedback pipelines \citep{bai2022constitutional,lee2024rlaif,yu2024rlaifv}. We do not, because
such a negative differs from the human positive in author, length and style as well as in fact, and
by \cref{prop:ratio} the reward learns every one of those differences; length alone is a known shortcut
for reward models \citep{singhal2024longway}. In our own test a head trained on human captions against
the policy's captions for the same images scores .36 on Multimodal RewardBench, below chance, and picks
the more hallucinated caption in every Best-of-$N$ cell (\cref{app:bon}). Editing one fact inside the same
response holds author, length and style fixed by construction.} These negatives cost nothing to label: the caption operators are string edits with no model in the
loop, and the 22k-pair bank builds in 10 seconds on one CPU core. Rendering the image arm's 1,212
pairs takes 5 GPU-hours (\cref{app:compute}).

\subsection{What the Objective Identifies}
\label{sec:identification}

We read \cref{eq:bt_loss} as ranking noise-contrastive estimation \citep{gutmann2010nce,ma2018rankingnce},
with the reward a linear energy over frozen features \citep{lecun2006ebm} and $\beta$ the InfoNCE
temperature \citep{oord2018cpc}. \edit{Write $d_{ij}=h^{+}_{i}-h^{-}_{j}$, so that \cref{eq:bt_loss} is $\frac{1}{B^{2}}\sum_{ij}\ell(\beta\,w^{\top}d_{ij})$
with $\ell(z)=\log(1+e^{-z})$. We keep three objects apart: an idealized objective over unrestricted scores (\cref{prop:ratio}); the linear
head under a fixed penalty $\frac{\lambda}{2}w^{\top}Dw$ (\cref{cor:direction}), a strongly convex problem
with one minimizer that the sampler and $\beta$ define and gradient descent reaches from any start; and the
heads AdamW trains, whose stationary points depend on its preconditioner \citep{xie2024adamw}, which we
measure (\cref{sec:plateau}). \Cref{app:identification} reads the two statements and \cref{app:theory} proves them.}

\begin{proposition}[the optimum is a density ratio against the sampler]
\label{prop:ratio}
\edit{Let $P$ and $Q$ be the joint laws of a positive and of a negative triple $(x,q,y)$, the latter with
the sampler of \cref{eq:sampler} averaged over the positive it edits, $\bar q(y\mid x,q)$, and assume
$P$ and $Q$ are mutually absolutely continuous. Over unrestricted scores, the cross-entry terms of
\cref{eq:bt_loss} are minimized exactly by}
\begin{equation}
\label{eq:ratio}
s^{\star}(x,q,y)\;=\;\frac{1}{\beta}\log\frac{\mathrm{d}P}{\mathrm{d}Q}(x,q,y)\;+\;C
\;=\;\frac{1}{\beta}\log\frac{p(y\mid x,q)}{\bar q(y\mid x,q)}\;+\;C ,
\end{equation}
\edit{with $C$ a constant. The $B$ same-entry terms, where $y^{-}$ is an edit of the very $y^{+}$ it is
scored against, are coupled and are not minimized by \cref{eq:ratio}.}
\end{proposition}

\edit{Those coupled terms carry weight $1/B=12.5\%$ at $B{=}8$ and move even the unrestricted optimum off
\cref{eq:ratio} (\cref{app:theory}). Our sampler also violates the absolute-continuity assumption, since a caption
of another image has probability zero for this one, so \cref{eq:ratio} describes what the cross-entry terms
reward and the linear head is analyzed through \cref{cor:direction}. Length drops out of the ratio when the
negatives match its law conditionally on content, and a matched marginal is not enough (\cref{app:identification});
our bank holds length fixed on two thirds of its pairs and shortens the negative by about 5\% on the rest, so
nothing makes the target length-free, and \cref{sec:honesty} finds the length artifact this allows.}

\paragraph{Why perturbations provide useful supervision.}
\edit{Independent negatives from the caption marginal, our $\mathcal{T}_{\times}$, make $s^{\star}$ the
image--caption pointwise mutual information up to a constant (\cref{app:pmi}): a score of topic
association, which can rank an accurate and an inaccurate response on the same topic equally. A
same-context edit changes one fact and nothing else. In the stylized model of \cref{app:cancel} the
quality factor cancels from the PMI score, while the edited pairs recover the correct ordering with
margin $\log\frac{\eta}{1-\eta}$ even when a fraction $1-\eta$ of the originals is wrong. Perturbations thus supply one quality comparison, an original above its degraded copy, without
requesting a judgment.}

\begin{corollary}[the sampler's pair law sets the direction]
\label{cor:direction}
With the penalty $\frac{\lambda}{2}w^{\top}Dw$, the population minimizer of \cref{eq:bt_loss} satisfies
\begin{equation}
\label{eq:representer}
\edit{D\,}w^{\star}\;=\;\frac{\beta}{\lambda}\;\mathbb{E}\big[\,\sigma(-\beta\,w^{\star\top}d)\;d\,\big],
\qquad d=h^{+}-h^{-},
\end{equation}
\edit{the expectation over the pair law of the matrix loss: $Dw^{\star}$ is the mean difference vector of the
sampler's pairs, each weighted by the probability the head still misorders it, so two samplers with the
same perturbations but different pair laws give different heads in general.}
\end{corollary}

\edit{The law of the differences, not their span, sets the direction; this is how the two CLIP-split heads of \cref{sec:dissociation} can reach
similar benchmark accuracy while differing by 6.4 points in hallucination judging. Nothing in the objective sets
the head's length coefficient to zero, and \cref{sec:honesty} finds it with opposite signs at 2B and 8B
under one sampler. Finally, \cref{eq:ratio} orders a policy's own samples by a ratio against the
perturbation process, so it ranks them by how unlike a perturbation each one looks and has no term
for how wrong it is. The design hypothesis this leaves is that a sampler drawn from the policy's own
errors ranks them better: \cref{sec:bon} measures how far the perturbation sampler falls short and
\cref{sec:midtraining} tests the hypothesis.}

\subsection{Experimental Setup}
\label{sec:experimental_setup}

We build pairs from 70k PixMo web captions \citep{deitke2024molmo} and train heads on Qwen3-VL
Instruct backbones at 2B, 4B, 8B and 32B, with the same negative bank and a frozen vision tower at
every size. The headline recipe is the 2B backbone at probe layer $N{-}4$, head only, 200 steps of 256 pairs
\edit{with AdamW at learning rate $10^{-3}$ and weight decay $0.01$}; every later run changes one element of it, and the curated comparison set is RLAIF-V
\citep{yu2024rlaifv}. Every head trains on the same schedule of 256 pairs per step, so two runs are
compared at equal pairs seen along the training timeline (\cref{fig:depth}, right).

We evaluate on Multimodal RewardBench \citep{yasunaga2025mmrewardbench}, VL-RewardBench
\citep{li2024vlrewardbench} and MMRB2 \citep{hu2025mmrb2}, with GPT-4o \citep{openai2024gpt4o} as the anchor on all three and
no re-tuning between them. Per-benchmark pair counts, the MMRB2 scoring construction and the
confidence bands are in \cref{app:protocol}.

For Best-of-$N$ we sample $N{=}16$ candidates from a Qwen3-VL-2B or 8B policy over 2,500 prompts of
COCO captioning \citep{lin2014coco}, MathVista \citep{lu2024mathvista}, MMMU \citep{yue2024mmmu}
and Claude-written hard questions on COCO images. We measure CHAIRs
\citep{rohrbach2018chair} or exact match against six training-free controls, from random to the
oracle, with 95\% confidence intervals from a paired bootstrap over prompts.

\section{Experiments}
\label{sec:exp}

\newcommand{\figdepth}{%
\begin{figure*}[t]
\centering
\begin{minipage}[t]{0.50\linewidth}\centering
\includegraphics[width=0.98\linewidth]{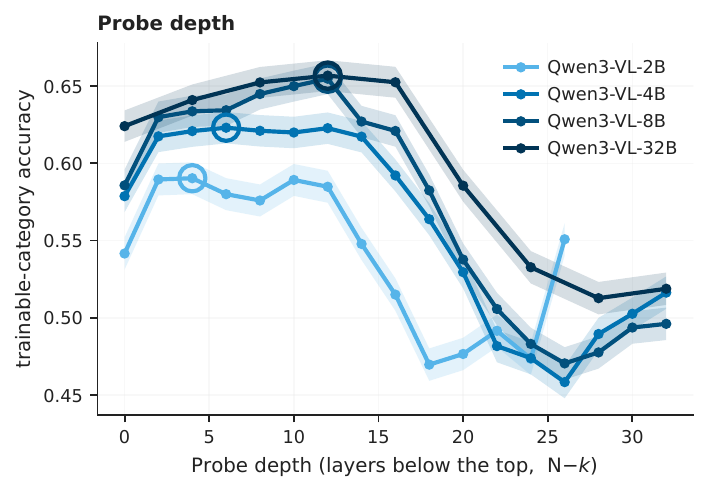}
\end{minipage}\hfill
\begin{minipage}[t]{0.44\linewidth}\centering
\includegraphics[width=0.96\linewidth]{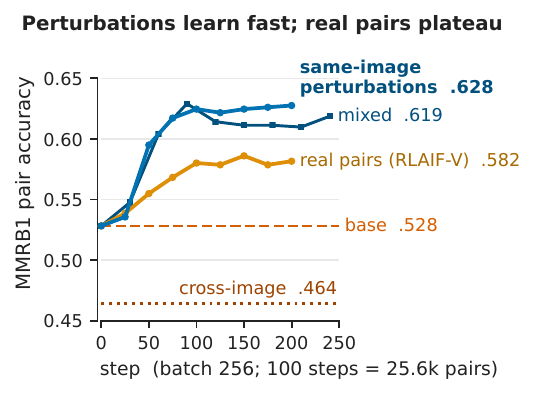}
\end{minipage}
\caption{\textbf{Left:} layer-sweep trainable accuracy against layers below the top for 2B / 4B / 8B / 32B
(bands: 75\% binomial intervals); the optimum is never the last layer, and sweep numbers rank depths within a
model only. \textbf{Right:} Multimodal RewardBench pair accuracy over the first 250 steps by negative
type (2B full fine-tune, 674-pair eval subset): same-image perturbations transfer, cross-image
negatives fall below the untrained base, curated pairs plateau lower.}
\label{fig:depth}
\vskip -0.1in
\end{figure*}}

\newcommand{\figscaling}{%
\begin{figure*}[t]
\centering
\includegraphics[width=0.98\linewidth]{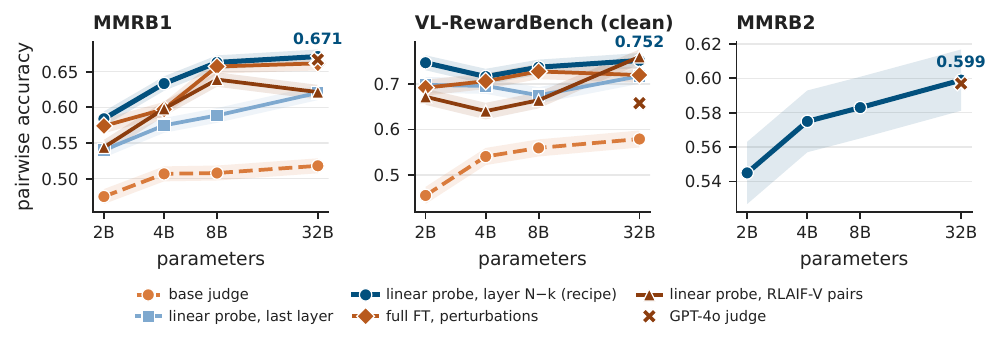}
\caption{Three reward benchmarks against backbone size, by component (shaded: confidence bands). The
linear probe at $N{-}k$ improves with size on all three and beats the last-layer probe at every size.
The full fine-tune matches it on MMRB1 only from 8B up and trails it on VL-RewardBench at 32B (early-stopped
at step 150 of 400). The RLAIF-V probe trails on MMRB1 everywhere and leads on VL-RewardBench only at 32B,
where the two share source images.}
\label{fig:scaling}
\vskip -0.1in
\end{figure*}}

\newcommand{\fighardqa}{%
\begin{figure*}[t]
\centering
\includegraphics[width=\linewidth]{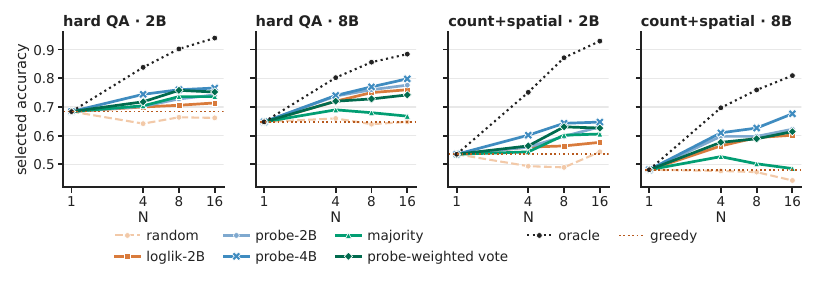}
\caption{Accuracy of the selected candidate against $N$ on 500 hard questions and the 241-question
counting/spatial subset, for the 2B and 8B generators, with every training-free control on every
panel. Where the generator is confidently wrong (8B) every external scorer beats majority vote; on the
2B pool probe, log-likelihood and vote are within noise. probe-8B omitted (it tracks probe-2B); its
numbers are in \cref{app:bon}.}
\label{fig:hardqa}
\vskip -0.1in
\end{figure*}}

\newcommand{\figloop}{%
\begin{figure*}[t]
\centering
\includegraphics[width=0.80\linewidth]{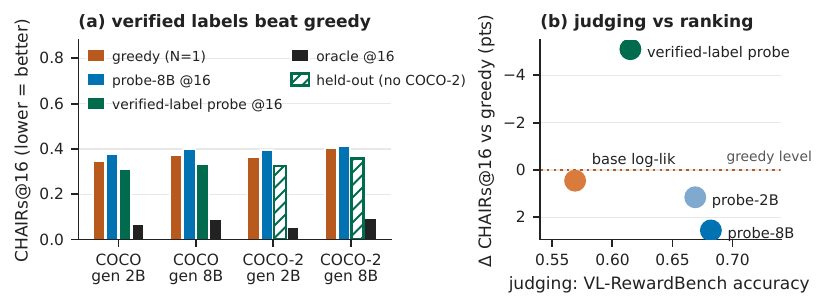}
\caption{One turn of the loop. \textbf{(a)} CHAIRs@16 per captioning cell (hatched: verified probe trained
without COCO-2). \textbf{(b)} Judging (VL-RewardBench) against ranking ($\Delta$CHAIRs vs greedy).}
\label{fig:loop}
\vskip -0.1in
\end{figure*}}

\newcommand{\figimageloop}{%
\begin{figure}[t]
\centering
\includegraphics[width=0.94\linewidth]{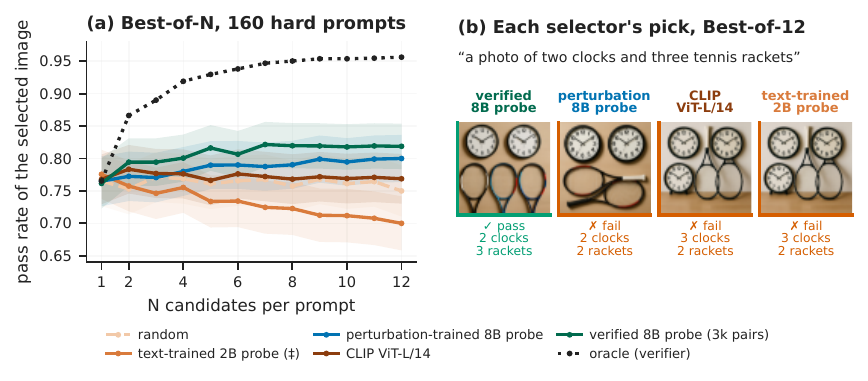}
\caption{\textbf{The loop transposes to image generation: verifier-minted pairs gain 5.4 points over
random selection at 8B, while the 2B head trained on caption perturbations loses 6.5.} \textbf{(a)} Best-of-$N$ on 160
held-out hard compositional prompts: verifier pass rate of the image each selector picks from $N$
Qwen-Image samples (bands: 75\% binomial intervals; paired differences against random at $N{=}12$ in the
text). \textbf{(b)} One prompt and each selector's pick out of twelve, judged by the GenEval-style OWLv2
verifier \citep{ghosh2023geneval,minderer2023owlv2} on object presence, count, colour and relative
position. The selectors see only the image.}
\label{fig:imageloop}
\vskip -0.1in
\end{figure}}


\subsection{Two Choices Set the Accuracy: The Probe Layer and the Sampler}
\label{sec:ablation}

Two choices set the accuracy, and neither is an optimizer setting: the probe layer alone adds 5.8
trainable points over the last layer, and mined negatives beat curated RLAIF-V pairs by 3.5 at equal
pair count. \Cref{tab:ablation4b} puts those two and every other ablation of the paper on one backbone under
one recipe. Every number is a real end-to-end run except the layer sweep that nominated the probe
depth: linear heads trained briefly on precomputed hidden states of every layer, a relative instrument
that appears only in \cref{fig:depth} and \cref{app:training}.
\figdepth

\paragraph{Probe depth.}
The best probe layer is never the top of the network: it is $N{-}4$ for 2B, $N{-}6$ for 4B and
$N{-}12$ for both 8B and 32B. That depth is chosen on Multimodal RewardBench trainable
accuracy, so we treat that benchmark as development throughout and VL-RewardBench and MMRB2, which no
choice in the recipe has seen, as the transfer evidence. The gap over the last layer is 3 to 7 points in the layer sweep
(\cref{fig:depth}) and 5.8 trainable points on the real 4B run. The optimum deepens from 2B to
8B and then holds at twelve layers below the top for 8B, 32B and GLM-4.1V-9B (36, 64 and 40 layers), while the
36-layer 4B peaks at $N{-}6$, so the depth is set by the backbone and not by its layer count; each peak is a plateau several layers
wide (\cref{app:training}).

\paragraph{Negative construction.}
We find that same-image perturbations transfer in both modalities and that cross-image negatives do not
(\cref{tab:corruption}). For text, $\mathcal{T}_{\times}$ alone reaches 99\% training accuracy and
falls below the untrained base on the benchmark, because it amplifies a length and style shortcut the
same-image operators mostly remove: that head scores .224 on the length-sensitive safety/bias column
against .451. For images the
replica is exact, with the deranged negative transferring nothing to MMRB2 text-to-image (52.7
against a label prior of 53.2) while the same-seed planted edit accounts for all of it (56.0). Perturbation density and the operator
mix are in \cref{app:training}.

\paragraph{Every training-side change leaves the benchmark within .014 of the recipe.}
\label{sec:plateau}
Changing the loss form or the negative composition, training on four times the pairs or for twice the steps,
or ordering the negatives from easy to hard or from hard to easy moves a
real 2B run by at most .014. In the sweep setting, changing the negative pool or the temperature schedule moves
it by less than .005. The sharpest row is a 540-configuration grid over head learning rate,
$\beta$ and weight decay: all 540 reach .995 or better in distribution, and selecting among them out
of distribution changes the benchmark by .0003.

\paragraph{Mined negatives against curated pairs at a matched pair budget.}
\label{sec:curated}
At 15k pairs the mined-negative head is 3 to 4 trainable points ahead of curated RLAIF-V pairs on
Multimodal RewardBench and 2 to 5 on VL-RewardBench, including the hallucination split
RLAIF-V was built for (\cref{tab:curated}). Unfreezing the backbone on the same 15k curated pairs
recovers 11 points overall to .613: a frozen head matches full fine-tuning on mined-negative data where
real pairs need the unfrozen model (\cref{app:training}).

\subsection{One Untuned Recipe Reaches GPT-4o Across Backbones and Benchmarks}
\label{sec:generalization}
\begin{table}[t]
\centering
\caption{\textbf{One recipe on three reward benchmarks.} Each probe is one frozen linear head at its
probe depth, trained once on caption perturbations and evaluated with no re-tuning. Published judges are
quoted from each benchmark's paper; our Multimodal RewardBench rows ($\dagger$) are over the 4,711
released pairs, which omit the withheld toxicity split (GPT-4o recomputed on them: .728 overall,
Gemini 1.5 Pro .734), and safety/bias is a length artifact for every mined-negative head
(\cref{sec:honesty}). VL-RewardBench macro is the mean of its three splits. The 2B and 8B rows use the
60k-pair bank, the 4B and 32B rows the 15k-pair rebuild (\cref{app:training}). Bold: column maximum
among our Qwen3-VL rows. \edit{The GLM-4.1V-9B rows are a second backbone family under the
identical recipe (15k pairs, 200 steps, $N{-}12$ of 40 layers, no re-tuning).}}
\label{tab:mmrb1}
\footnotesize
\setlength{\tabcolsep}{3pt}
\begin{tabular}{l c cc c cc c c}
\toprule
\textbf{Model} & \textbf{Overall} & \multicolumn{2}{c}{\textbf{General}} & \textbf{Knowl.}
  & \multicolumn{2}{c}{\textbf{Reasoning}} & \textbf{Safety} & \textbf{VQA} \\
\cmidrule(lr){3-4}\cmidrule(lr){6-7}
 & & Corr. & Pref. & & Math & Coding & Bias & \\
\midrule
Claude 3.5 Sonnet & .720 & .626 & .678 & .739 & .686 & .651 & .768 & .856 \\
Gemini 1.5 Pro    & .720 & .635 & .677 & .663 & .689 & .555 & .945 & .872 \\
GPT-4o            & .715 & .626 & .690 & .720 & .676 & .621 & .748 & .872 \\
\midrule
Ours: 2B probe, $N{-}4$   & .708$^\dagger$ & .669 & .610 & .587 & .654 & .490 & \textbf{.998} & .850 \\
Ours: 4B probe, $N{-}6$   & .724$^\dagger$ & .713 & .621 & .624 & .716 & .500 & .951 & .855 \\
Ours: 8B probe, $N{-}12$  & .659$^\dagger$ & \textbf{.735} & \textbf{.638} & .635 & .708 & \textbf{.579} & .187 & .860 \\
Ours: 32B probe, $N{-}12$ & \textbf{.733}$^\dagger$ & .726 & .636 & \textbf{.643} & \textbf{.739} & .567 & .742 & \textbf{.911} \\
\edit{Ours: GLM-4.1V-9B probe, $N{-}12$} & \edit{.692$^\dagger$} & \edit{.636} & \edit{.602} & \edit{.514} & \edit{.562} & \edit{.612} & \edit{.935} & \edit{.857} \\
\midrule
base 4B (log-likelihood judge) & .571$^\dagger$ & .449 & .558 & .581 & .582 & .364 & .191 & .892 \\
\edit{base GLM-4.1V-9B (log-likelihood judge)} & \edit{.565$^\dagger$} & \edit{.440} & \edit{.495} & \edit{.573} & \edit{.611} & \edit{.404} & \edit{.167} & \edit{.892} \\
\bottomrule
\end{tabular}\\[0.7em]
\begin{tabular}{lccccc}
\toprule
\textbf{VL-RewardBench} & \textbf{Overall} & \textbf{Macro} & \textbf{General} & \textbf{Halluc.} & \textbf{Reason.} \\
\midrule
Gemini-1.5-Pro (strongest published) & .672 & -- & -- & -- & -- \\
GPT-4o (published)                   & .658 & -- & -- & -- & -- \\
\midrule
Ours: 32B probe, $N{-}12$ & \textbf{.724} & \textbf{.649} & \textbf{.475} & \textbf{.810} & \textbf{.663} \\
Ours: 8B probe, $N{-}12$  & .682 & .607 & .425 & .765 & .631 \\
Ours: 4B probe, $N{-}6$   & .680 & .600 & .420 & .774 & .606 \\
Ours: 2B probe, $N{-}4$   & .669 & .592 & .425 & .762 & .587 \\
\edit{Ours: GLM-4.1V-9B probe, $N{-}12$} & \edit{.602} & \edit{.531} & \edit{.348} & \edit{.678} & \edit{.568} \\
\midrule
base 8B (log-likelihood judge) & .542 & .525 & .459 & .554 & .561 \\
base 4B (log-likelihood judge) & .525 & .518 & .464 & .522 & .568 \\
base 2B (log-likelihood judge) & .569 & .550 & .525 & .598 & .527 \\
\edit{base GLM-4.1V-9B (log-likelihood judge)} & \edit{.570} & \edit{.529} & \edit{.425} & \edit{.616} & \edit{.546} \\
\bottomrule
\end{tabular}
\vskip -0.1in
\end{table}

The 32B probe is at or above GPT-4o on all three benchmarks and
separably above it on one (\cref{tab:mmrb1}). On VL-RewardBench it
is the strongest entry on the table, 6.6 points above GPT-4o at 3.6 standard errors of the difference;
on Multimodal RewardBench and on MMRB2 reasoning the benchmark cannot resolve the margin, and
separating either at two standard errors would take 13 and 3.5 times the released pairs. The stronger
reference than GPT-4o is the backbone itself as a generative judge, which on MMRB2, the one benchmark
we ran it on, leads the probe by 4.7 points \edit{(64.6 against 59.9 on the four-task average, \cref{app:protocol})}. In exchange, the probe decodes nothing: it needs one forward pass per
response, where the judge generates a judgment in both presentation orders. Larger backbones improve the intermediate-layer probe on all three benchmarks,
while the base log-likelihood judge stays near chance and the last-layer head trails at every size
(\cref{fig:scaling}).
\edit{The recipe is not specific to Qwen3-VL: on GLM-4.1V-9B, a second family with 40 layers, the same
recipe at $N{-}12$ beats its last layer and its untrained judge on both benchmarks (\cref{tab:mmrb1}).}
A head trained only on caption perturbations also transfers a usable signal to response-image tokens
once the backbone is large enough (MMRB2 editing, level with GPT-4o), which starts \cref{sec:imageloop}. Our training data shares no prompt with the benchmarks and 0.1\% of images with Multimodal RewardBench
(\cref{app:repro}).

\paragraph{What the columns mean.}
\label{sec:honesty}
The safety/bias column of Multimodal RewardBench is a length artifact, the one reward models are
known to carry \citep{singhal2024longway}. A head trained on
same-image perturbations, two thirds of which match length exactly, scores .998 on it at 2B and .19 at 8B, where its correlation with
response length flips sign. The gap is not seed noise: eight layer-sweep seeds give $.967\pm.007$ at 2B and $.569\pm.013$ at
8B, the same ordering (\cref{app:protocol}). We therefore compare heads on the five \emph{trainable} categories, where the 8B probe's .659 overall
against .708 at 2B becomes .658 against .602.

\subsection{Two Probes, One Score, Two Rewards}
\label{sec:dissociation}

\paragraph{Same benchmark score, different reward.}
We train two probes on the two extreme quartiles of the same mined-negative data, ranked by the drop in
CLIP \citep{radford2021clip} image--text cosine from the real to the perturbed caption, 15k
pairs each at 2B. They reach similar accuracy on Multimodal RewardBench, .597 against .603 trainable, and
then separate three ways (\cref{fig:teaser}b): VL-RewardBench hallucination 72.0 against 78.4,
reasoning 63.4 against 56.8, and as Best-of-$N$ scorers the CLIP-visible head beats probe-8B on
CHAIRs by 3.8 points while losing 2.9 on QA. The same pattern, a small benchmark gap against a
large hallucination gap, holds across sizes, consistent with \cref{cor:direction}: the two samplers
differ from the positives along different coordinates, and the benchmark does not separate them. 

\paragraph{Whether a scorer helps Best-of-$N$ depends on the pool, not on its benchmark score.}
\label{sec:bon}
\edit{Across eight captioning and QA cells and a hard-QA ladder, a probe's benchmark accuracy does not
predict how much it improves Best-of-$N$ selection, in either direction (\cref{tab:hardqa,app:bon}).} On captioning every mined-negative probe stays at greedy level where the oracle reaches .06 to
.09 CHAIRs, and on MathVista and MMMU, where candidate agreement tracks correctness, majority vote is
1.4 to 3.8 points ahead of the probe, whose one positive role there is as a vote weighter. Where the 8B generator is confidently wrong instead
(majority wrong on 32\% of prompts), every external scorer beats majority vote: probe-2B by 10.8
points, probe-8B by 7.2 and the training-free log-likelihood judge by 9.2, all with 95\% CIs
excluding zero. The trained head does not significantly beat the training-free judge: probe-2B against log-likelihood is $+1.6$ to
$+2.1$ points with CIs including zero, and probe-2B beats probe-8B by 3.6 points, the reverse of
their benchmark order. \edit{What decides the gain is the pool: whether the generator's majority can be trusted, and so whether
any external scorer should replace the vote. The benchmark measures neither.}

\subsection{Learning from the Policy's Own Errors: Better Deployment, Lower Benchmarks}
\label{sec:midtraining}
\figloop

\edit{\Cref{fig:method}b changes one element of the recipe of \cref{sec:algorithm}, the negative
sampler: the policy's own Best-of-$N$ candidates, labelled by a cheap verifier, are paired correct
against wrong for the same prompt. We train the head to separate verified successes from failures among
the policy's own outputs, the comparison that Best-of-$N$ selection makes at deployment.}

\edit{It lowers captioning CHAIRs where the perturbation probes did not:} a head trained on 3,462 such pairs cuts
CHAIRs by 5.1 points against greedy and by 4.1 points on a task it never
trained on (\cref{fig:teaser}c, \cref{fig:loop}a, \cref{tab:loop}). On QA it matches probe-8B as a selector, so the utility claim
rests on captioning, the open-ended task where no vote is available. CHAIR is both harvest label and metric, so the gain is in absent COCO-object
mentions rather than faithfulness at large (\cref{app:limitations}).

\edit{The verified head is the weakest judge of the
three heads (\cref{fig:loop}b) on VL-RewardBench, .615 against .669 and .682 for the perturbation probes, yet the one
that ranks the policy's captions below greedy's hallucination rate (the other benchmarks in
\cref{tab:loop}). Both follow from the sampler: 70\% of the verified pairs are QA correctness pairs, so
the categories they do not cover fall while math rises. The two rewards differ because the two
comparisons do. Benchmark pairs contrast other models' responses on annotator-chosen errors, close to
what perturbations produce; verified pairs contrast this policy's own successes and failures. In our runs the head trained on
the second kind improves candidate selection and loses VL-RewardBench accuracy. Serving both the benchmark and the deployment is open: train on the union of the two samplers, give the head more
than one direction, or draw benchmark pairs from the policy under deployment.}

\paragraph{The actor side of the turn.}
\edit{One DPO turn on pairs the policy generated itself lowers its COCO hallucination rate from .446 to
.390 CHAIRs with probe-picked pairs, .414 with verified and .412 with
random pairs, at no cost on MathVista or MMMU (\cref{tab:actor_dpo}).} The label adds little over random pairs, so the selector's
effect on the actor is still to be shown (\cref{sec:appendix_actor}).

\figimageloop
\paragraph{The same turn on image generation.}
\label{sec:imageloop}
The loop transposes to image generation point for point (\cref{fig:imageloop}):
Qwen-Image \citep{qwenimage2025} produces pools of 12 on GenEval-style prompts \citep{ghosh2023geneval}
and a label-free OWLv2 verifier \citep{minderer2023owlv2} plays CHAIR's role. On 160 held-out hard
prompts (random .765, oracle .956), the text-trained 2B head is 6.5 points below random (95\% CI
$-11.6$ to $-1.6$), and the mined-negative image heads are $+2.5$, on par with CLIP ($+3.0$). A head
trained on 3,000 verifier-minted pairs is $+5.4$ at 8B (95\% CI $+0.5$ to $+10.3$), 2.4 points above
CLIP and 2.9 above the mined-negative image heads on the same prompts. The two verified
heads cross over as MMRB2 judges, so here too the better ranker is the worse judge.

\section{Related Work}
\label{sec:related}

\textbf{Multimodal reward models and the benchmarks that rank them.} A multimodal reward model is
built by annotating preference pairs and fitting a scorer on them, from LLaVA-RLHF and RLHF-V
\citep{sun2023llavarlhf,yu2024rlhfv} to MM-RLHF's 120k human comparisons \citep{zhang2025mmrlhf} and
the scoring heads of IXC-2.5-Reward and UnifiedReward \citep{zang2025ixcreward,wang2025unifiedreward}.
Removing the annotator usually means a model in their place \citep{bai2022constitutional,lee2024rlaif,\edit{li2024vlfeedback,}yu2024rlaifv}; we use a sampler\edit{, as hallucination-injection DPO does for the policy \citep{zhou2024povid,zhao2023hadpo}}.
Multimodal RewardBench \citep{yasunaga2025mmrewardbench}, VL-RewardBench \citep{li2024vlrewardbench}
and MMRB2 \citep{hu2025mmrb2} rank those models and generative judges \citep{zheng2023judge\edit{,xiong2025llavacritic}} by
pairwise accuracy against human labels \citep[cf.][]{lambert2024rewardbench\edit{,malik2026rewardbench2,frick2025ppe}}. We take them as given and ask what the number measures. \edit{In text, policies optimized toward reward models of similar accuracy can perform quite differently \citep{wen2025rethinking}, while an outcome reward model's score tracks downstream accuracy better than validation loss does \citep{qi2026evolm}; we ask when a reward model's own benchmark score stops tracking deployment.}

\textbf{Probing and probe depth.} The scorer we set against those judges is a probe, and
linear probes are classically an analysis tool for encoders \citep{alain2017probing,tenney2019bert,burns2023dlk}\edit{; intermediate layers often give better features than the last \citep{skean2025layer}}.
LLaVA-Reward \citep{zhou2025llavareward} is the closest prior use of a probe as a reward, mixing
depths through a cross-attention module trained on annotated preferences; ours is a 2,048-parameter head with no label.

\textbf{Density-ratio estimation and negative sampling.} A probe on a frozen backbone approximates a density ratio, which noise-contrastive estimation fits
against a noise distribution
\citep{gutmann2010nce,oord2018cpc} that the negative-sampler literature designs to shape the representation \citep{he2020moco,robinson2021hard,kalantidis2020mochi,wang2020alignuniform}\edit{, and in vision-language models as word-order and attribute swaps of the true caption \citep{yuksekgonul2023bow,hsieh2023sugarcrepe}}. With frozen features those choices act only through the objective. Closest to our objective is \citet{fan2026rbs}, a label-free
in-batch Bradley--Terry reward on text with web continuations as the negative process; their
invariance theorem gives a nuisance's coefficient in the unrestricted optimum, which \cref{prop:ratio}
restates for our matrix loss.

\textbf{Best-of-$N$ and learning from self-generated data.} A reward built from that ratio is then used, like a verifier \citep{cobbe2021verifiers\edit{,uesato2022process,lightman2024verify}}, under optimization pressure, and proxy rewards degrade under it \citep{gao2023overopt,\edit{coste2024ensembles,}stiennon2020summarize}. We locate an earlier failure, before any pressure is
applied, where the proxy's benchmark score does not predict its Best-of-$N$ value \edit{\citep{beirami2025bon}} against
self-consistency \citep{wang2023selfconsistency}. Expert iteration and STaR
\citep{anthony2017exit,zelikman2022star\edit{,singh2024restem}} improve a policy on its own verified outputs, and GKD
\citep{agarwal2024gkd} diagnoses the train--inference mismatch on the policy side; ours is the
reward-side analogue.

\section{Concluding Remarks}
\label{sec:conclusions}

Frozen vision-language representations support strong reward modeling with only a linear head and
caption perturbations, but probe depth and training comparisons determine which signals the head
extracts. Similar benchmark scores can conceal different specializations, and higher preference
accuracy need not yield better Best-of-$N$ selection. Retraining on verified policy outputs improves
caption selection while lowering benchmark accuracy, so the supervision that helps deployment is drawn
from the policy's own errors. Anyone choosing a reward model should test how well it picks the best of
their own policy's samples, and a paper that reports one should state which pairs it was trained on.

\section*{Acknowledgement}
S.K. acknowledges the support from the National Science Foundation Grant under award IIS 2229881; H.Z. and S.K. acknowledge the Chan Zuckerberg Initiative Foundation for establishing the Kempner Institute for the Study of Natural and Artificial Intelligence.

\bibliography{ref}
\bibliographystyle{plainnat}

\clearpage
\appendix
\renewcommand{\contentsname}{Appendix Contents}
\addtocontents{toc}{\protect\setcounter{tocdepth}{2}}
\tableofcontents
\clearpage

\newcommand{\tabdpo}{%
\begin{table}[H]
\centering
\caption{\textbf{Actor, DPO (2B policy).} Full-parameter DPO ($\beta{=}0.1$, lr $10^{-6}$, 2 epochs,
vision tower frozen) on three corpora from the same 500 prompts $\times$ 4 tasks: \emph{verified}
(3,462 pairs), \emph{rm} (probe-8B argmax vs argmin, 2,000), \emph{random} (two random candidates,
1,999; the control). Greedy regeneration at 512 tokens; $\Delta$ against the regenerated base, paired
bootstrap over prompts (2,000 resamples), every QA CI including zero. CHAIRi is hallucinated object mentions
over all object mentions, pooled over the 500 COCO captions, which shorter captions do not lower by themselves.
Best value per column in bold.}
\label{tab:actor_dpo}
\footnotesize
\setlength{\tabcolsep}{5pt}
\begin{tabular}{lcccc}
\toprule
 & \multicolumn{2}{c}{COCO} & MathVista & MMMU \\
\cmidrule(lr){2-3}\cmidrule(lr){4-4}\cmidrule(lr){5-5}
\textbf{setting} & words & CHAIRi $\downarrow$ & acc ($\Delta$) & acc ($\Delta$) \\
\midrule
base (regenerated)        & 245 & .083 & \textbf{.546} & .334 \\
verified (3,462 pairs)    & 242 & \textbf{.073} & .542 ($-.004$) & \textbf{.346} ($+.012$) \\
rm (2,000)                & 228 & .079 & .528 ($-.018$) & .338 ($+.004$) \\
random (1,999, control)   & 245 & .082 & .534 ($-.012$) & .332 ($-.002$) \\
\bottomrule
\end{tabular}
\end{table}}

\newcommand{\tabdensity}{%
\begin{table}[H]
\centering
\caption{\textbf{Perturbation density,} $k{=}1$ / 3 / all editable spans per rejected caption (all $=$ 13.6
edits on average), 21,521 identical images and captions per setting, token substitutions only. Frozen 2B
$N{-}4$ head, 200 steps of 1,024, single seed. The training task gets much easier (val .82 $\to$ .996)
and neither benchmark moves.}
\label{tab:density}
\footnotesize
\setlength{\tabcolsep}{4pt}
\begin{tabular}{lcccccc}
\toprule
\textbf{setting} & val & MMRB1 & \multicolumn{4}{c}{VL-RewardBench} \\
\cmidrule(lr){4-7}
 & & overall & overall & macro & general & halluc. \\
\midrule
$k{=}1$ (1.0 edit)      & .820 & .673 & .709 & .665 & .630 & .781 \\
$k{=}3$ (3.0 edits)     & .973 & .673 & \textbf{.713} & .660 & .613 & .797 \\
$k{=}$all (13.6 edits)  & .996 & \textbf{.681} & .707 & .644 & .547 & .797 \\
\midrule
ref: mixed-operator probe & -- & .682 & .679 & -- & -- & .769 \\
\bottomrule
\end{tabular}
\end{table}}

\newcommand{\tabdecoys}{%
\begin{table}[H]
\centering
\caption{\textbf{Planted decoys in the captioning pools.} 3.4 single-error decoys written by an LLM and
4 template perturbations are planted among the 16 real candidates of each COCO prompt. AUROC of real over
decoy, the rate at which the argmax falls on a decoy (chance .31), and CHAIRs of the pick with and
without decoys. Every scorer ranks a source above its own one-edit decoy $\ge 93\%$ of the time, by only
0.3 to 1.1 within-pool $\sigma$; log-likelihood's ``detection'' is self-style (it picks a 2B-style decoy
on 88\% of 8B-pool prompts).}
\label{tab:decoys}
\footnotesize
\setlength{\tabcolsep}{4pt}
\resizebox{\linewidth}{!}{%
\begin{tabular}{llcccc}
\toprule
\textbf{pool} & \textbf{scorer} & AUROC, LLM decoys & AUROC, template & decoy pick & CHAIRs of pick (full / orig.) \\
\midrule
COCO 2B & probe-2B  & .609 & .866 & .010 & .374 / .368 \\
COCO 2B & probe-8B  & .643 & .876 & .004 & .362 / .358 \\
COCO 2B & loglik-2B & .700 & .917 & .000 & .348 / .348 \\
COCO 8B & probe-2B  & .802 & .888 & .074 & .428 / .384 \\
COCO 8B & probe-8B  & .884 & .889 & .030 & .406 / .384 \\
COCO 8B & loglik-2B & \textbf{.084} & .912 & \textbf{.884} & .914 / .382 \\
\bottomrule
\end{tabular}}
\end{table}}

\section{Limitations and Broader Impacts}
\label{app:limitations}
Our approach lowers the additional training and preference-annotation requirements for multimodal
reward modeling by learning a small scoring head over a frozen backbone. This may make task-specific
reward models more accessible to researchers and practitioners with limited annotation budgets, while
retaining the inference cost of the underlying backbone. Constructing training comparisons explicitly
also makes the intended supervision easier to inspect and revise.

The broader implication concerns how reward models are evaluated and selected. Our experiments show
that higher preference-benchmark accuracy need not yield better selection of a policy's outputs.
Reporting the training comparison distribution alongside deployment-specific evaluations can therefore
support more informed model selection. Conversely, using benchmark performance alone to justify
deployment risks overlooking systematic errors and shortcuts, including response-length biases.

These benefits have a limited demonstrated scope. Our experiments cover two backbone families, and
verifier-based adaptation is evaluated for one round. Captioning improvements are measured with CHAIR,
which also supplies the training labels; they establish fewer absent COCO-object mentions rather than
general factuality or preserved informativeness, and a CHAIR-independent metric such as POPE
\citep{li2023pope} would be needed to settle the latter. Best-of-$N$ cells are single-seed, the hard-QA
set was written and verified by one LLM family, and our Multimodal RewardBench overall omits the
withheld toxicity split (\cref{sec:honesty}). The method does not establish safety or fairness
capabilities, and broader deployment requires independent evaluation of output quality, coverage, and
task-specific failure modes.

\section{Reproducibility}
\label{app:repro}
The recipe is fully specified by \cref{eq:readout,eq:bt_loss,eq:sampler} and the hyper-parameters of
\cref{sec:experimental_setup} and \cref{app:training}: a frozen Qwen3-VL checkpoint, a probe layer,
mean pooling over response tokens, and one bias-free linear head trained with AdamW for 200 steps.
Every input is public. Positives are PixMo-Cap captions \citep{deitke2024molmo}; negatives are string
edits of those captions with no model in the loop, so the 22k-pair bank rebuilds in 10 seconds on one
CPU core from the operator definitions in \cref{app:training}. The curated comparison set is public
RLAIF-V \citep{yu2024rlaifv}, and all three evaluation benchmarks
\citep{yasunaga2025mmrewardbench,li2024vlrewardbench,hu2025mmrb2} are used as released, with the pair
counts and the scoring construction stated per benchmark in \cref{sec:experimental_setup}. Contamination is
checked by a 256-bit difference hash of every training image against every evaluation image, counting
a pair as a near-duplicate at $\le 10$ differing bits, plus exact prompt matching. Against Multimodal
RewardBench the perturbation set shares 5 near-duplicate images of 4,711 (0.1\%) and no prompt; against
VL-RewardBench, the 1,000 MMRB2 prompt images and the 1,000 COCO Best-of-$N$ images it shares none.
The image arm renders 1,212 PixMo-derived prompts with Qwen-Image, so its training images are
generated and its prompts are not those of MMRB2. The same audit exposes the curated baseline:
RLAIF-V pairs overlap VL-RewardBench on 326 of 1,247 rows and Multimodal RewardBench on 18, which is
why every RLAIF-V-trained row also reports the decontaminated subset. Source for pair construction, training, every selector in the Best-of-$N$
tables and the plotting scripts behind every figure will be released with the paper.

\section{Evaluation Protocol}
\label{app:protocol}
Multimodal RewardBench releases 4,711 of its 5,211 pairs, the toxicity split being withheld, in eight
categories; our overall number is therefore comparable to published ones only through the
recomputation of \cref{tab:mmrb1}. VL-RewardBench has 1,247 pairs (general 181, hallucination 749,
reasoning 317), of which 326 overlap the RLAIF-V training set, so we also score anything trained on RLAIF-V
on the 921 clean rows. On the 326 overlapping rows the RLAIF-V fine-tunes score .80 against .60 to
.64 for models that never saw the data; on the clean rows their lead disappears. MMRB2 has 1,000 judge-labelled pairs per task (text-to-image, editing,
interleaved generation, reasoning; 996 reasoning pairs are scoreable). We score it with the prompt in
the user turn and the response, images included, in the assistant turn, the construction the heads
were trained on. A scalar reward is order-invariant, so its accuracy equals MMRB2's dual-order
accuracy; exact ties count 0.5. We also report reasoning on the 350-pair text-only-output
subset, the convention that benchmark's paper uses for scalar reward models. \edit{On the four-task
average under this construction the 32B probe scores 59.9 and the same backbone used as a generative
judge 64.6; GPT-4o as a judge scores 59.7.} Scoring prompt and
response in one user turn instead costs the text heads 2 to 3 points on reasoning and 5 to 9 on its
text-only subset. The GPT-4o margins in \cref{tab:mmrb1} are quoted as unpaired standard errors because the benchmark
papers do not release per-item predictions, so a paired interval is not available. Shaded bands in every figure are 75\% binomial intervals of the plotted accuracy
over its evaluation set; seed-to-seed spread of the heads is narrower than the line width. Eight seeds of the layer-sweep heads put the safety/bias column of \cref{sec:honesty} at
$.967\pm.007$ at 2B and $.569\pm.013$ at 8B, where the correlation between head score and response
length is $-0.57$ at 2B and $+0.44$ at 8B. The Best-of-$N$ intervals of \cref{sec:bon} are over prompts at one selector seed.

The Best-of-$N$ prompts are 500 each of COCO captioning (two prompt sets, COCO and COCO-2), MathVista
and MMMU, plus 500 Claude-written hard questions on COCO images, of which 241 are counting or
spatial; candidates are sampled at temperature 1.0 and top-$p$ 0.95, plus the greedy decode. The
mid-training runs of \cref{sec:midtraining} continue from the mined-negative probe on 3,462 verified
pairs harvested from those pools (MathVista 1,352, MMMU 1,072, COCO 511, COCO-2 527), with
leave-one-task-out runs that drop every COCO-2 or every MMMU pair, and a non-convex check that
unfreezes the backbone on 15k RLAIF-V pairs from either checkpoint. Heads train with AdamW, learning
rate $10^{-3}$, weight decay $0.01$, $\beta{=}0.1$, micro-batch 8 and effective batch 256 or 1,024 as
stated; the backbones have 28, 36, 36 and 64 layers.

\section{Negative Mining and Probe Depth}
\label{app:training}
Two pair banks appear in the benchmark tables: the 2B and 8B rows of \cref{tab:mmrb1} use the original
60k-pair bank, the 4B and 32B rows and every ablation the rebuilt 15k-pair bank, which costs the 2B
probe 2 to 3 points (.708 to .682 on Multimodal RewardBench), so the 32B row leads from the smaller
bank. Perturbation density does not move either benchmark: one, three or all editable spans stay within .008 on both
benchmarks (\cref{tab:density}) while the training task gets much easier. The three token-edit heads
beat the mixed-operator recipe on VL-RewardBench by 2.8 to 3.4 points at equal Multimodal RewardBench
accuracy (\cref{tab:corruption}). That comparison also drops the cross-image and the length-changing
operators, so it does not isolate the mix.
\begin{table}[t]
\centering
\caption{\textbf{Every ablation and analysis of the paper on one backbone (Qwen3-VL-4B).} Frozen
linear head, mean pooling, probe on layer $N{-}6$ unless stated, 200 steps of 256 pairs, single seed,
no re-tuning between benchmarks. MMRB1 = Multimodal RewardBench (4,711 pairs; \emph{trainable} = the
five categories without safety/bias and VQA); VL-RB = VL-RewardBench (1,247 pairs, hallucination split
749). The $\Delta$-CLIP heads are 5k-pair quartiles of the same perturbations split by whether CLIP can
see the edit; the verified heads train on 2,400 verifier-minted pairs from a fresh head or from the
mined-negative head; the base judge scores by response log-likelihood with no training.}
\label{tab:ablation4b}
\footnotesize
\setlength{\tabcolsep}{3pt}
\begin{tabular}{lcccc}
\toprule
\textbf{setting} & \textbf{MMRB1 overall} & \textbf{MMRB1 trainable} & \textbf{VL-RB} & \textbf{VL-RB halluc.} \\
\midrule
mined-negative probe, $N{-}6$ (recipe)      & \textbf{.724} & \textbf{.633} & .680 & \textbf{.774} \\
\quad last-layer control                & .696 & .575 & .659 & .700 \\
\quad frozen head on RLAIF-V 15k        & .676 & .598 & .618 & .677 \\
\quad $\Delta$-CLIP blind / visible     & .687 / .707 & .615 / .631 & .656 / .673 & .721 / .765 \\
\quad verified, fresh / mid-training    & .705 / .723 & .615 / .632 & .616 / \textbf{.690} & .650 / .762 \\
\midrule
base log-likelihood judge               & .571 & .507 & .525 & .522 \\
\bottomrule
\end{tabular}
\vskip -0.1in
\end{table}

\figcorruption
\begin{table}[t]
\centering
\caption{\textbf{Negative-mining ablation in both modalities.} Top: text responses, 2B backbone. The
frozen-head rows use 15k pairs at $N{-}4$ on the official benchmarks; the two full-fine-tune rows are
the June series that first separated the negative types (overall Multimodal RewardBench; their
safety/bias column reads .224 and .451, the length shortcut). Bottom: image responses, 2B image-trained
head at $N{-}12$ on 1,091 same-seed prompt pairs; \emph{held-out} is accuracy on the head's own negative
type, MMRB2 text-to-image is judge agreement (label prior 53.2). The \emph{by kind} row splits held-out
accuracy of the combined head by the planted edit.}
\label{tab:corruption}
\footnotesize
\setlength{\tabcolsep}{4pt}
\begin{tabular}{llccccc}
\toprule
\textbf{text negatives} $y^-$ & \textbf{head} & \textbf{val} & \textbf{MMRB1} & \textbf{trainable} & \textbf{VL-RB} & \textbf{halluc.} \\
\midrule
cross-image only, $\mathcal{T}_{\times}$              & full FT     & .99  & .559 & --   & --   & -- \\
same-image perturbations, mixed                         & full FT     & --   & .582 & --   & --   & -- \\
token edit, $k{=}1$ span, $\mathcal{T}_{\mathrm{near}}$ & frozen head & .820 & .673 & .574 & .709 & .781 \\
token edit, $k{=}3$ spans                              & frozen head & .973 & .673 & .563 & \textbf{.713} & \textbf{.797} \\
token edit, all spans (13.6)                           & frozen head & .996 & .681 & .576 & .707 & \textbf{.797} \\
mixed operators, 15\% $\mathcal{T}_{\times}$ (recipe)  & frozen head & .92  & \textbf{.682} & \textbf{.584} & .679 & .769 \\
\bottomrule
\end{tabular}\\[0.6em]
\begin{tabular}{llcc}
\toprule
\textbf{image negatives} $y^-$ & & \textbf{held-out, own type} & \textbf{MMRB2 T2I} \\
\midrule
cross-image only, $G(q';s')$                          & & .93--1.00 & 52.7 \\
planted edit only, $G(\mathcal{T}(q);s)$              & & .79--.85  & 56.0 \\
both (recipe)                                          & & .79--.84  & \textbf{56.2} \\
\quad by kind: object swap / near-miss / multi-edit    & 2B & .944 / .714 / .837 & \\
\quad                                                  & 8B & .917 / .786 / .884 & \\
\bottomrule
\end{tabular}
\vskip -0.1in
\end{table}

\tabdensity
\begin{table}[t]
\centering
\caption{\textbf{Mined negatives against curated RLAIF-V pairs at a matched pair budget,} 2B backbone,
15k pairs per setting. Frozen heads: $N{-}4$, 200 steps of 1,024 pairs. Full fine-tune: lr $10^{-5}$, 400
steps of 256, vision tower frozen. The \emph{tokens} column is the response-token count of the 15k pairs (both
sides; Qwen3-VL tokenizer): perturbation captions are longer than RLAIF-V responses (204 against 86
tokens per side), so the RLAIF-V heads see fewer tokens at the same pair count. The \emph{val} column is
in-distribution pair accuracy on each setting's own held-out pairs. 326 of the 1,247 VL-RewardBench rows
overlap the RLAIF-V training set, so RLAIF-V models are also read on the 921 clean rows; mined-negative pairs
overlap none. Single seed; the comparison is matched in pairs and not in tokens.}
\label{tab:curated}
\footnotesize
\setlength{\tabcolsep}{4pt}
\resizebox{\linewidth}{!}{%
\begin{tabular}{llcccccc}
\toprule
\textbf{training pairs (15k)} & \textbf{head} & \textbf{tokens} & \textbf{val} & \textbf{MMRB1} & \textbf{trainable} & \textbf{VL-RB} & \textbf{VL-RB clean} \\
\midrule
perturbations, token-edit      & frozen head & 6.4M & .820 & .673 & .574 & .709 & \textbf{.747} \\
perturbations, mixed operators & frozen head & 6.1M & .92  & \textbf{.682} & \textbf{.584} & .679 & -- \\
RLAIF-V                      & frozen head & 2.6M & .551 & .577 & .544 & .660 & .672 \\
RLAIF-V                      & full FT     & 2.6M & .637 & \textbf{.686} & \textbf{.613} & \textbf{.735} & .727 \\
\midrule
base 2B (log-likelihood judge) & --        & --   & --   & .563 & --   & .569 & -- \\
\bottomrule
\end{tabular}}
\vskip -0.1in
\end{table}

\Cref{fig:depth} gives the depth sweep per model, \cref{tab:corruption} the operator ablation in both
modalities, \cref{tab:density} the density variants and \cref{tab:curated} the matched-budget
comparison against RLAIF-V. On that matched budget the mined-negative head reads .574 to .584
trainable on Multimodal RewardBench against .544 for RLAIF-V and .679 to .709 on VL-RewardBench
against .660, with hallucination .769 to .781 against .737; its in-distribution validation accuracy
is .82 to .92 against .55 for the curated pairs. The comparison is matched in pairs and not in
tokens, since perturbation captions run 2.4 times longer, which leaves the RLAIF-V head 2.6M response
tokens against 6.1M. Unfreezing the backbone on the same 15k real pairs gains 11 points overall and 7
trainable, to .613, which puts it above the mined-negative probe on the trainable categories and
level with it on decontaminated VL-RewardBench (.727 against .747). Perturbation operators act on one caption at a time: a near-miss substitution of a colour, number or
spatial word; a splice of another caption's tail; a clause shuffle; and, for 15\% of pairs, the caption
of another image. \edit{In the 15k-pair bank the operators and their share are near-miss substitution 50\%, attribute
substitution 15\%, splice 17\%, shuffle 10\% and cross-image 8\%. Measured as the ratio of rejected to
chosen length in words, the two substitutions leave length unchanged (1.00), splice shortens to 0.82,
shuffle to 0.69 (it drops about 30\% of sentences) and cross-image captions are 1.10 times as long,
so across the bank a negative is about 5\% shorter than its positive.} \edit{On GLM-4.1V-9B the same recipe
at the last layer scores .628 on Multimodal RewardBench and .598 on VL-RewardBench against .692 and .602
at $N{-}12$ (\cref{tab:mmrb1}); the untrained judge scores .565 and .570.} Density variants perturb one, three or all editable spans (13.6 on average) of the
same 21,521 captions. Heads train with AdamW, learning rate $10^{-3}$, weight decay $0.01$,
$\beta{=}0.1$, micro-batch 8, effective batch 256 or 1,024 as stated, 200 steps, on precomputed vision-tower
outputs; the 32B confirm run used FSDP \citep{zhao2024fsdp2} over eight GPUs with 1,024 effective. The layer sweep trains a linear head on the precomputed hidden states of each layer for 8k pairs and
five seeds; it ranks depths within a model and nothing else, and it misleads in two documented ways.
Last-token pooling ranks first in the sweep at 4B, 8B and 32B and loses 7.5 points to mean pooling in
the one real run. Sweep heads trained under a different regime than the real pipeline disagree with it
by several points, in either direction.

\section{Extended Results for Reward-Model Training}
\label{app:dynamics}
\figscaling
\Cref{fig:scaling} separates the components behind \cref{tab:mmrb1} at each backbone size.

\section{Extended Results for \texorpdfstring{Best-of-$N$}{Best-of-N}}
\label{app:bon}
\fighardqa
\begin{table}[t]
\centering
\caption{\textbf{Best-of-16 on hard verifiable QA} (500 questions on COCO images, and the 241-question
counting/spatial subset), accuracy of the selected candidate for each selector. \emph{greedy} is
$N{=}1$; \emph{oracle} is pass@16. Paired bootstrap over prompts; single seed per selector. Where the
8B generator is confidently wrong (mean top-share .85) every external scorer beats majority vote. On
the 2B pool, the probes and log-likelihood are within noise of each other and of the vote.}
\label{tab:hardqa}
\footnotesize
\setlength{\tabcolsep}{3.5pt}
\resizebox{\linewidth}{!}{%
\begin{tabular}{lccccccccc}
\toprule
\textbf{cell} & greedy & random & shortest & loglik-2B & probe-2B & probe-8B & majority & wvote & oracle \\
\midrule
hard QA, gen 2B        & .684 & .672 & .580 & .714 & .742 & \textbf{.762} & .736 & .752 & .940 \\
hard QA, gen 8B        & .648 & .648 & .448 & .760 & \textbf{.776} & .740 & .668 & .742 & .884 \\
counting/spatial, gen 2B & .535 & .502 & .440 & .577 & \textbf{.631} & .622 & .606 & .627 & .929 \\
counting/spatial, gen 8B & .481 & .436 & .253 & .602 & \textbf{.622} & .610 & .485 & .614 & .809 \\
\bottomrule
\end{tabular}}
\vskip -0.1in
\end{table}

\begin{table}[t]
\centering
\caption{\textbf{One turn of the loop, 2B.} Top: CHAIRs at $N{=}16$ per captioning cell (lower is
better); \emph{verified} is the head trained on 3,462 verifier-minted pairs, \emph{held-out} the same
recipe with every COCO-2 pair excluded, scored on COCO-2. Pooled paired bootstrap against greedy:
$-5.1$ CHAIR points, 95\% CI $-7.1$ to $-3.1$; held-out $-4.1$, 95\% CI $-6.9$ to $-1.1$. Bottom: the same heads as
judges on the two unseen reward benchmarks, against the mined-negative probe they replace (the 15k-pair
rebuild that scores every Best-of-$N$ cell). The best ranker
is the worst recent judge.}
\label{tab:loop}
\small
\setlength{\tabcolsep}{4.5pt}
\begin{tabular}{lccccc}
\toprule
\textbf{captioning cell} & greedy & probe-8B & verified & held-out & oracle \\
\midrule
COCO, gen 2B   & .346 & .376 & \textbf{.312} & -- & .066 \\
COCO, gen 8B   & .374 & .400 & \textbf{.332} & -- & .090 \\
COCO-2, gen 2B & .362 & .396 & \textbf{.294} & .324 & .056 \\
COCO-2, gen 8B & .402 & .414 & \textbf{.344} & .358 & .094 \\
\bottomrule
\end{tabular}\\[0.6em]
\begin{tabular}{lccc}
\toprule
\textbf{judge} & VL-RB & halluc. & MMRB2 reas. \\
\midrule
verified, fresh init      & .615 & 68.8 & .539 \\
verified, perturbation init & .615 & 68.9 & .540 \\
mined-negative probe & \textbf{.679} & \textbf{76.9} & .533 \\
\bottomrule
\end{tabular}
\vskip -0.1in
\end{table}

\Cref{fig:hardqa} and \cref{tab:hardqa} give the hard-QA ladder with every control on every panel and
\cref{tab:loop} the loop turn per captioning cell. Initializing the verified head from the mined-negative probe matters only
when the pair budget is too small to reach the optimum. At 2B a fresh head and a continued one score
within 0.3 points of each other everywhere (\cref{tab:loop}); at 4B, where 2,400 pairs do not reach the optimum,
the continued head is 7 points ahead (\cref{tab:ablation4b}). A planted-decoy test (3.4 LLM-written single-error decoys and 4 template perturbations per
captioning pool) shows that the mined-negative feature fires inside a real pool: every scorer ranks a source above
its own one-edit decoy at least 93\% of the time. The margin, however, is 0.3 to 1.1 within-pool standard deviations, and
CHAIRs of the pick does not move. The probe therefore detects that an edited sentence is worse than its source
but cannot tell which of 16 real captions is the most truthful. Truth-preserving edits separate the signals: the probes do not prefer
the source over a true edit (probe-2B .512, probe-8B .291) while log-likelihood prefers it 91\% of the time.

\paragraph{Model-written negatives.}
\edit{Two 2B heads were trained with the policy's own captions in the pair instead of a string edit,
same recipe otherwise. Self-corruption (chosen: the policy's caption; rejected: its perturbation) ties
greedy and random in all four captioning cells and scores .689 on Multimodal RewardBench, the same as
the off-policy head. Human-versus-policy (chosen: the PixMo caption; rejected: the policy's caption for
the same image) scores .360 on Multimodal RewardBench, below chance, and picks a more hallucinated
caption than random in all four cells (CHAIRs@16 .422 / .432 / .414 / .422 against random .370 / .384 /
.406 / .402). This head learned to tell a human author from a model; on a benchmark where both responses
are model-written, that direction anti-correlates with the label.}

\section{Actor Training}
\label{sec:appendix_actor}
\label{app:actor}
We train the 2B policy on its own selected candidates against a size-matched random-selection
control. DPO \citep{rafailov2023dpo} uses $\beta{=}0.1$, learning rate $10^{-6}$, two epochs, and three
corpora from the same 500 prompts per task: verifier-minted pairs, probe-8B argmax-versus-argmin
pairs, and random pairs as the control. Policies are re-evaluated greedily on all four tasks.

The actor half of the turn shows no selector effect, though it has a safe recipe. DPO costs nothing on QA:
no change is outside $\pm1.8$ points, and the per-object hallucination rate on COCO falls for every
corpus (CHAIRi .083 to .073, .079 and .082 for the verified, RM-picked and random pairs), with the label
adding little over random pairs (\cref{tab:actor_dpo}). The size-matched random control plays
the role of the untrained-initialization control of \citet{fan2026rbs}. The RM-picked corpus is the only one
that shortens captions (7\%), yet its per-object rate also falls, so the change is not purely length. One
DPO turn on self-generated pairs is a mild, label-agnostic hallucination reducer, and the actor side of the
loop still lacks a demonstrated selector effect.
\tabdpo
The DPO base policy was regenerated at 512 new tokens under the identical protocol. Between 19 and 27\% of DPO
pairs were skipped by the 1,024-token cap, mostly long QA chains of thought, so the QA share of each
corpus is below nominal; the update was small (loss .69 to .64 to .68). The image-generation
counterpart (LoRA \citep{hu2022lora} on Qwen-Image \citep{qwenimage2025} for 160 steps on 640
on-policy images, across seven Diffusion-DPO \citep{wallace2024diffusiondpo} and related fine-tuning variants) has the same shape. No variant differs from the base generator on the hard
prompts (all $\Delta$ within $[-.04,+.07]$), and every selection variant minus its random control is
inside $\pm2$ points.

\section{CLIP Studies}
\label{app:clip}
\begin{figure}[H]
\centering
\includegraphics[width=\linewidth]{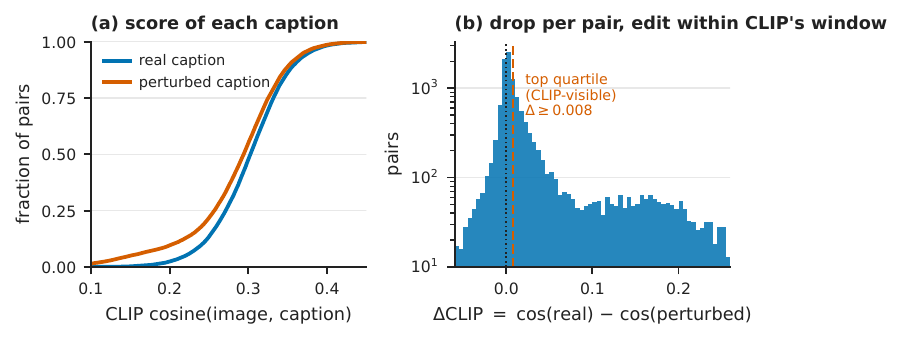}
\caption{\textbf{CLIP view of the perturbation pairs} (22,000 same-image pairs, CLIP ViT-B/32). (a) Image--caption cosine of
the real and the perturbed caption; the perturbation lowers the median from .303 to .294 and adds a low-score tail.
(b) The drop $\Delta$CLIP per pair, which the dissociation split of \cref{sec:dissociation} ranks on, on a log count
axis; the dashed line is the top-quartile cut ($\Delta\geq.008$) that defines the CLIP-visible arm. CLIP accepts 77
tokens and the captions average 150 words, so for 43\% of pairs the edit falls beyond its window and $\Delta=0$
exactly; the CLIP-blind quartile is drawn from those pairs and is omitted from the histogram.}
\label{fig:clipcdf}
\end{figure}
\Cref{fig:clipcdf} shows the score distribution the quartile cuts are taken from.
Filtering the positives by image--caption CLIP score gives .685 / .698 / .678 on Multimodal
RewardBench for the top, random and bottom quartile and .670 / .686 / .682 on VL-RewardBench. By
contrast, the $\Delta$CLIP split ranks each pair by the drop in CLIP cosine from the real to the perturbed
caption. The CLIP-blind quartile has near-zero drop, part of it because CLIP truncates at 77 tokens
and the edit falls outside the window.

\section{Reading the Two Statements}
\label{app:identification}
\edit{The convexity fact of \cref{sec:identification}} predicts that, when gradient descent runs on a fixed-penalty objective, initialization, step size and step count cannot move the
minimizer. It predicts nothing about AdamW, whose stationary points depend on its preconditioner, and nothing about the negative pool, the hardness schedule or the temperature, each
of which changes the population objective rather than the path to its minimizer. The temperature is
the least obvious of these: writing $u=\beta w$ turns the weight decay into
$\frac{\lambda}{2\beta^{2}}\lVert u\rVert^{2}$, so $\beta$ acts as an effective regularization
strength that rescales $w^{\star}$ and tilts its direction. That those interventions nonetheless leave
the benchmark where it was (\cref{sec:plateau}) is an empirical finding, and \cref{eq:representer}
explains it. Each intervention re-weights the same difference vectors, and on the benchmark's pairs the
re-weighted heads order alike. The CLIP split of \cref{sec:dissociation} is the same equation evaluated at
a re-weighting large enough to separate on hallucination. The value at the minimizer is the best
linear separation of positives from negatives in the layer $N{-}k$ feature space, which is consistent with depth
moving the plateau while optimizer settings do not; convexity fixes the endpoint of the fixed-penalty problem and not the
distance to it, which is where initialization re-enters (\cref{sec:midtraining}).

\Cref{prop:ratio} is stated for the decoupled objective because that is the one with a closed-form
optimum, and for mutually absolutely continuous $P$ and $Q$ because that is what keeps the log ratio
finite. Our sampler does not satisfy the hypothesis: a caption of another image has probability zero
as a caption of this one, so on those pairs the unrestricted optimum runs to $\pm\infty$ and separates
the two laws with unbounded margin. The in-distribution accuracy of the recipe is the finite-sample
face of that, with all 540 configurations of \cref{sec:plateau} at .995 or better, and it is why we
analyze the linear head through the fixed-penalty problem of \cref{cor:direction} rather than through \cref{eq:ratio}. The diagonal terms alone would leave a context-dependent offset free, making scores
comparable within a prompt and not across prompts; the matrix form removes that freedom, and our
same-image-only ablation
(\cref{sec:plateau}) confirms that the offset does not change within-prompt accuracy. The matrix form
also decides what the target does with length. Write a response as content $c$ and length $l$. If the negatives match the positives' length law
conditionally on content, $p(l\mid c)=\bar q(l\mid c)$, then $\mathrm{d}P/\mathrm{d}Q$ depends on $c$ alone and length
drops out of \cref{eq:ratio} (cf.\ \citealp{fan2026rbs}). A matched marginal is not enough: with two contents
and two lengths, the joint tables
\[
  P=\begin{pmatrix}.4&.1\\.1&.4\end{pmatrix},\qquad Q=\begin{pmatrix}.1&.4\\.4&.1\end{pmatrix}
\]
share the length marginal $(.5,.5)$ and still have ratios $4$ and $1/4$ that depend on length. Our
substitution edits hold length fixed pairwise and splice and shuffle shorten the negative, so the bank meets
neither condition exactly and nothing makes the target of \cref{eq:ratio} length-free. A linear head cannot
represent \cref{eq:ratio} exactly and is characterized by the error-weighted mean difference of \cref{eq:representer} instead, whose
coefficient on length is not the target's; the sign flip of \cref{sec:honesty} is that head-side
coefficient changing with the backbone under one sampler.

\subsection{What Constructed Comparisons Supervise}
\label{app:comparisons}
\edit{Fix a context $x$ (an image and its prompt) and a finite response set $\mathcal{Y}$, and let
$J_{x}(y,y')$ be the probability that a training comparison designates $y$ as chosen and $y'$ as
rejected, so that the conditional population loss is
\begin{equation}
\label{eq:jloss}
\mathcal{L}_{x}(s)\;=\;\sum_{y,y'\in\mathcal{Y}} J_{x}(y,y')\,\ell\big(s(x,y)-s(x,y')\big),
\qquad \ell(v)=\log(1+e^{-v}).
\end{equation}
If $\mu(y\mid x)$ is the distribution of real captions and $K_{x}(y'\mid y)$ constructs the negative,
then $J_{x}(y,y')=\mu(y\mid x)K_{x}(y'\mid y)$: independent negatives use $K_{x}(y'\mid y)=\nu(y')$,
our cross-image operator $\mathcal{T}_{\times}$, and a deterministic edit uses
$K_{x}(y'\mid y)=\mathbf{1}\{y'=\mathcal{C}(y)\}$. Judged candidates give a $J_{x}$ in which winner and
loser are dependent after labeling, so their marginals alone do not determine the objective; the joint
comparison law does. These are population statements about unrestricted scalar scores and carry no
finite-sample or downstream guarantee.}

\subsection{Independent Negatives: Density Ratios and PMI}
\label{app:pmi}
\edit{With $J_{x}(y,y')=\mu(y\mid x)\,\nu(y')$ and a common support $S_{x}$, the argument of
\cref{prop:ratio} with the context held fixed gives the unrestricted minimizers of \cref{eq:jloss} on
$S_{x}$,
\begin{equation}
\label{eq:indep}
s^{\ast}(x,y)\;=\;\log\frac{\mu(y\mid x)}{\nu(y)}+\phi(x),\qquad y\in S_{x},
\end{equation}
with $\phi(x)$ arbitrary. If $\nu=\mu_{Y}$, the caption marginal of the corpus law $\mu(x,y)$, then
\begin{equation}
\label{eq:pmi}
s^{\ast}(x,y)\;=\;\log\frac{\mu(x,y)}{\mu_{X}(x)\,\mu_{Y}(y)}+\phi(x)\;=\;\mathrm{PMI}_{\mu}(x,y)+\phi(x),
\end{equation}
the idealized case of using other images' captions as negatives. PMI measures image--caption
association, not correctness or preference agreement; shared support is an assumption a finite corpus
does not guarantee; and the result does not apply to edited copies or to judge-ordered candidates.}

\subsection{When Perturbations Expose an Additional Quality Signal}
\label{app:cancel}
\edit{The following stylized example separates topic association from response quality. Represent a
response as $y=(t,c)$, where $t$ denotes its topic and $c\in\{0,1\}$ indicates an incorrect or correct
version. Suppose the corpus has distribution
\begin{equation}
\label{eq:topicmodel}
\mu\big((t,c)\mid x\big)=p_{x}(t)\,b_{\eta}(c),\qquad b_{\eta}(1)=\eta,\quad b_{\eta}(0)=1-\eta,\quad
\tfrac12<\eta<1.
\end{equation}
Thus topics depend on the context, but the fraction of correct versions is the same across contexts and
topics. Let $p(t)$ be the corpus topic marginal and assume $p_{x}(t)>0$ on a common finite topic set.
Independent corpus negatives have distribution $\mu_{Y}(t,c)=p(t)b_{\eta}(c)$, so \cref{eq:indep} gives
\begin{equation}
\label{eq:indep_topic}
s^{\ast}_{\mathrm{ind}}(x,t,c)\;=\;\log\frac{p_{x}(t)\,b_{\eta}(c)}{p(t)\,b_{\eta}(c)}+\phi(x)
\;=\;\log\frac{p_{x}(t)}{p(t)}+\phi(x).
\end{equation}
The quality factor cancels: the optimum distinguishes topics by their association with the context but
assigns equal scores to correct and incorrect versions within a topic.

Now consider an idealized edit $\mathcal{C}(t,c)=(t,1-c)$ that switches the version while preserving
the topic, with the original always labeled chosen. This label agrees with correctness with
probability $\eta$ and is reversed with probability $1-\eta$; the corpus need not be perfectly
correct. For a fixed topic, write $\Delta_{t}=s(x,t,1)-s(x,t,0)$. Its conditional loss and minimizer
are
\begin{equation}
\label{eq:editloss}
\mathcal{L}_{\mathrm{edit},t}(\Delta_{t})=\eta\,\ell(\Delta_{t})+(1-\eta)\,\ell(-\Delta_{t}),\qquad
\Delta_{t}^{\ast}=\log\frac{\eta}{1-\eta}>0.
\end{equation}
Indeed, the derivative is $\sigma(\Delta_{t})-\eta$, which vanishes uniquely at the stated margin.
Perturbation training therefore ranks the correct version above the incorrect one, a distinction absent
from the independent-negative optimum in this example.

A Bradley--Terry judge with reward $r(x,t,c)=\gamma c$, $\gamma>0$, gives the same
within-topic ordering with margin $\gamma$; the edited pairs recover the ordering with a margin set by
$\eta$ instead. Under the stated factorization and edit assumptions, the example shows one additional quality
distinction, not a general superiority of perturbations. Changing quality-bearing content inside
an otherwise matched response can supply a preference that corpus association alone does not teach;
the experiments test whether actual edit rules do so and whether that preference transfers.}

\section{Proofs}
\label{app:theory}

\paragraph{Convexity of the regularized objective.}
Write the population objective as
\[
  L(w)\;=\;\mathbb{E}\,\ell\big(\beta\,w^{\top}d\big),
  \qquad d=h^{+}-h^{-},
  \qquad \ell(z)=\log\big(1+e^{-z}\big).
\]
The function $\ell$ is twice differentiable with
\[
  \ell''(z)\;=\;\sigma(z)\,\sigma(-z)\;>\;0 ,
\]
so it is strictly convex, and a strictly convex function of an affine map is convex.

The Hessian of $L$ is
\[
  \nabla^{2}L(w)\;=\;\beta^{2}\,\mathbb{E}\Big[\ell''\big(\beta w^{\top}d\big)\,d\,d^{\top}\Big].
\]
It is positive definite on $\operatorname{span}\{d\}$, because the weights $\ell''$ are strictly
positive, and zero on the orthogonal complement, because $w^{\top}d$ does not depend on that
component. Hence $L$ is strictly convex on the span and constant on its complement.

Adding the penalty turns the objective into
\[
  L_{\lambda}(w)\;=\;L(w)\;+\;\frac{\lambda}{2}\,w^{\top}Dw,
  \qquad \lambda>0,\; D\succ0 ,
\]
which is $\lambda\,\lambda_{\min}(D)$-strongly convex on all of $\mathbb{R}^{m}$, with $m$ the feature dimension. It therefore has exactly one minimizer.
If $\mathbb{E}\lVert d\rVert^{2}<\infty$, it is also $L_{s}$-smooth with
$L_{s}=\tfrac{\beta^{2}}{4}\mathbb{E}\lVert d\rVert^{2}+\lambda\lambda_{\max}(D)$, since $\ell''\le\tfrac14$, and gradient
descent with a constant step below $2/L_{s}$ converges to the minimizer from every initialization.
Isotropic weight decay is $D=I$.

\edit{AdamW \citep{loshchilov2019adamw} is not gradient descent on $L_{\lambda}$ for a fixed $D$. With a frozen diagonal
preconditioner $M$, its update $w\leftarrow w-\alpha(M\nabla L(w)+\lambda w)$ stops where
\[
  \nabla L(w)+\lambda M^{-1}w=0 ,
\]
which moves with $M$: for $L(w)=\log(1+e^{-w})$ and $\lambda=1$ the fixed point solves $w=M\sigma(-w)$, giving
$0.40$ at $M=1$ and $1.04$ at $M=4$. The actual $M$ follows the gradient history, and \citet{xie2024adamw}
characterize AdamW's limit points as KKT points of an $\ell_\infty$-constrained problem instead. The statements
in this appendix therefore concern the fixed-penalty problem; the AdamW heads of the experiments are
measured, not derived (\cref{sec:plateau}).}

For gradient descent, initialization, step size, sample order and the number of steps change only the iterate sequence, so
at fixed $p$, $q_{\mathrm{neg}}$, $\beta$, $\lambda$ and $D$ none of them changes the minimizer.

The negative pool, its hardness and its composition change $q_{\mathrm{neg}}$ and therefore the law
of $d$, so they may move the minimizer; \cref{eq:representer} says how, through the error-weighted
mean of $d$. The temperature is not neutral either. Writing
$u=\beta w$,
\[
  L_{\lambda}(w)\;=\;\mathbb{E}\,\ell\big(u^{\top}d\big)\;+\;\frac{\lambda}{2\beta^{2}}\,\|u\|^{2},
\]
so $\beta$ enters as an effective regularization strength $\lambda/\beta^{2}$ and a change in it
rescales $w^{\star}$ and, since ridge shrinkage is isotropic while $\mathbb{E}[dd^{\top}]$ is not,
tilts its direction as well. Only the accuracy of a \emph{ranking} is invariant to the scale of
$w^{\star}$, not to its direction. \qed

\paragraph{Proof of \cref{prop:ratio}.}
Let $P=\mu\otimes p$ and $Q=\mu\otimes\bar q$ be the joint laws of a positive and of a negative
triple $a=(x,q,y)$, with $\bar q$ the marginal law of a negative as in the statement. The
off-diagonal population loss over unrestricted scores $s$ is
\[
  J(s)\;=\;\mathbb{E}_{a\sim P,\;b\sim Q}\;\ell\big(\beta\,(s(a)-s(b))\big),
\]
where $a$ and $b$ are independent because they come from different micro-batch entries. We assume
throughout that $P$ and $Q$ are mutually absolutely continuous, so that $\mathrm{d}P/\mathrm{d}Q$ is
finite and positive almost everywhere. The hypothesis is what makes the minimizer a real-valued
function: on a pair where one of the two densities vanishes, the stationarity condition below puts
$\delta^{\star}$ at $\pm\infty$, the infimum of $J$ over real-valued scores is not attained, and every
sequence approaching it separates the two laws with unbounded margin. Our own sampler is of that
kind, which is why the recipe's weight decay and its linear class, and not \cref{eq:ratio}, are what
pin down the head that gets trained (\cref{cor:direction}).

Group the contributions of each unordered pair $\{a,b\}$ of triples. The pair appears once with $a$
as the positive, at density $P(a)Q(b)$, and once with $b$ as the positive, at density $P(b)Q(a)$.
For a fixed difference $\delta=s(a)-s(b)$ its contribution is
\[
  P(a)Q(b)\,\ell(\beta\delta)\;+\;P(b)Q(a)\,\ell(-\beta\delta).
\]
This is strictly convex in $\delta$, and its derivative vanishes where
\[
  \sigma(\beta\delta)\;=\;\frac{P(a)Q(b)}{P(a)Q(b)+P(b)Q(a)} ,
  \qquad\text{that is, at}\qquad
  \beta\,\delta^{\star}\;=\;\log\frac{P(a)}{Q(a)}\;-\;\log\frac{P(b)}{Q(b)} .
\]

The function $s^{\star}=\beta^{-1}\log(\mathrm{d}P/\mathrm{d}Q)+C$ attains $\delta^{\star}$ for every
pair simultaneously, so it minimizes every summand and therefore $J$. Any other minimizer attains
$\delta^{\star}$ on almost every pair as well, because each summand is strictly convex in $\delta$,
so it differs from $s^{\star}$ by a constant on the support of $P$ and $Q$. Since the two laws share
the context law $\mu$,
\[
  \frac{\mathrm{d}P}{\mathrm{d}Q}(x,q,y)
  \;=\;\frac{\mu(x,q)\,p(y\mid x,q)}{\mu(x,q)\,\bar q(y\mid x,q)}
  \;=\;\frac{p(y\mid x,q)}{\bar q(y\mid x,q)} ,
\]
which is \cref{eq:ratio}. The pairs run over triples and not over responses within one context, so
no context-dependent offset is left free; only the constant $C$ is. The factor $\beta^{-1}$ is where
the temperature enters the target, and a linear head absorbs it into the scale of $w$.

The $B$ diagonal terms are outside this argument. There $b$ is an edit of the same $a$ it is scored
against, so the pair law is $P(a)\,q_{\mathrm{neg}}(b\mid a)$ rather than the product $P(a)Q(b)$, the
two orderings of $\{a,b\}$ no longer enter at weights that depend only on the marginals, and
$\delta^{\star}$ above is not the pointwise minimizer. We therefore state the proposition for the
decoupled objective, whose terms carry weight $1-1/B=87.5\%$ at $B{=}8$.

The coupling moves the optimum. Take two responses with $p=(.8,.2)$, an edit that always swaps them, so
$\bar q=(.2,.8)$, and $\beta=1$. Up to a constant from equal-response pairs, the full objective at $B{=}8$ is
\[
  \tfrac78\big(.64\,\ell(\Delta)+.04\,\ell(-\Delta)\big)+\tfrac18\big(.8\,\ell(\Delta)+.2\,\ell(-\Delta)\big)
  \;=\;.66\,\ell(\Delta)+.06\,\ell(-\Delta),
\]
minimized at $\Delta=\log 11$, while the cross-entry terms alone give $\log 16$. \qed

\paragraph{Proof of \cref{cor:direction}.}
The population objective with the penalty is
\[
  L_{\lambda}(w)\;=\;\mathbb{E}\,\ell\big(\beta\,w^{\top}d\big)\;+\;\frac{\lambda}{2}\,w^{\top}Dw,
\]
with the expectation over the pair law of the matrix loss: each micro-batch contributes $B$
same-entry pairs and $B^{2}-B$ cross-entry pairs, at total weights $1/B$ and $1-1/B$. Since
$\ell'(z)=-\sigma(-z)$,
\[
  \nabla L_{\lambda}(w)\;=\;-\beta\,\mathbb{E}\big[\sigma(-\beta\,w^{\top}d)\,d\big]\;+\;\lambda\,Dw ,
\]
and $L_{\lambda}$ is strongly convex \edit{(previous paragraph)}, so its unique minimizer is the unique zero
of this gradient, which is \cref{eq:representer}. The inner product of $Dw^{\star}$ with a unit vector $v$ gives its
coefficient along $v$. In particular $Dw^{\star}$ is a positive combination of difference vectors and lies in their span, which is all of $\mathbb{R}^{m}$ whenever the difference vectors span it, and then constrains
nothing; what \cref{eq:representer} fixes is the combination.

Now take $D=I$ and suppose the features encode length along a unit direction $v$ with $v^{\top}d=0$ almost
surely on the same-entry pairs. This is an assumption on the backbone, which pairwise length preservation
(exact for the substitution operators, approximate for the bank) makes plausible and does not imply. Those
pairs then contribute nothing to the head's coefficient $v^{\top}w^{\star}$, and the cross-entry pairs, which compare captions of different images, contribute
\[
  \frac{\beta}{\lambda}\Big(1-\frac{1}{B}\Big)\,
  \mathbb{E}_{\times}\big[\sigma(-\beta\,w^{\star\top}d)\,v^{\top}d\big],
\]
an error-weighted mean of gaps between unrelated captions, which the objective does not set to zero.
Pairwise preservation on the diagonal therefore leaves the coefficient along $v$ to the cross-entry
pairs rather than fixing it at zero, the same conclusion as for the unrestricted optimum above, where
a comparison that holds a nuisance fixed leaves its coefficient undetermined. For $D\neq I$ the same computation
gives $v^{\top}Dw^{\star}$, and the head's coefficient $v^{\top}w^{\star}$ can be nonzero even when every difference is
orthogonal to $v$: with $D=\mathrm{diag}(1,2)$ and $d=(1,-1)$, $w^{\star}\propto(1,-\tfrac12)$ and
$v=(1,1)/\sqrt2$ gives $v^{\top}d=0$ and $v^{\top}w^{\star}>0$.

For two samplers $\bar q_{1},\bar q_{2}$ with difference laws $\rho_{1}\neq \rho_{2}$, \cref{eq:representer} gives
minimizers $w_{1},w_{2}$ that coincide only if the two error-weighted means do, a condition on the
laws and not on their spans. A benchmark pair set $\{e_{k}\}$ measures the sign pattern
\[
  \operatorname{sign}\big(w^{\top}e_{k}\big),\qquad k=1,\dots,K ,
\]
which coincides for $w_{1}$ and $w_{2}$ whenever the two heads order the benchmark's pairs alike,
however far apart they are on other pairs. \qed

\section{Compute Resources}
\label{app:compute}
Training a head is the cheapest step in the study. Once the frozen backbone's hidden states are
stored, a 540-configuration grid at 2B fits in about nine GPU-minutes, and a single head is 200 steps
of a $d\to1$ linear map. The 32B confirm run used FSDP over eight GPUs at effective batch 1,024. The
compute goes to everything that is not the reward model: one forward pass of the backbone per pair to store
its hidden states, Best-of-$N$ sampling at $N{=}16$ over 2,500 prompts for two policy sizes, and the image arm's
renders. The image arm needs two same-seed Qwen-Image renders per prompt at about 7 GPU-seconds each,
about 5 GPU-hours for its 1,212 training pairs. The caption negative bank itself is free: 10 seconds on one CPU core for
22k pairs, with no model in the loop. The figure the paper quotes against fine-tuning, about a
thousandth of the compute, counts head training against the full fine-tune of the same backbone on the
same pairs and excludes the backbone forward passes both settings share.

\end{document}